\documentclass[acmsmall]{acmart}

\usepackage{booktabs} 

\usepackage{mathtools}
\usepackage{amsmath}

\setcopyright{rightsretained}

\begin{document}
\title[Copula-based anomaly scoring and localization of high-dimensional data]{Copula-based anomaly scoring and localization for large-scale, high-dimensional continuous data}

\author{G\'abor Horv\'ath}
\affiliation{%
  \institution{Dept. of Networked Systems and Services, Budapest University of Technology and Economics}
  \streetaddress{Magyar Tud\'osok krt. 2}
  \city{Budapest}
  \country{Hungary}
  \postcode{1117}
}
\email{ghorvath@hit.bme.hu}

\author{Edith Kov\'acs}
\affiliation{%
  \institution{Dept. of Differential Equations, Budapest University of Technology and Economics \\
  Faculty of Informatics, University of Debrecen}
  \streetaddress{M\H{u}egyetem rkp. 3.}
  \city{Budapest}
  \country{Hungary}
  \postcode{1111}
}
\email{kovacsea@math.bme.hu}

\author{Roland Molontay}
\orcid{0000-0002-0666-5279}
\affiliation{%
  \institution{Dept. of Stochastics, Budapest University of Technology and Economics, \\ MTA-BME Stochastics Research Group \\
  Faculty of Informatics, University of Debrecen}
  \streetaddress{P.O. Box 91}
  \city{Budapest}
  \country{Hungary}
  \postcode{1521}
  }
\email{molontay@math.bme.hu}

\author{Szabolcs Nov\'aczki}
\affiliation{%
  \institution{Nokia, Bell Labs}
  \streetaddress{B\'okay J\'anos utca 36-42}
  \city{Budapest}
  \country{Hungary}
  \postcode{1083}
}
\email{szabolcs.novaczki@nokia-bell-labs.com}

\begin{abstract}
The anomaly detection method presented by this paper has a special feature: it does not only indicate whether an observation is anomalous or not but also tells what exactly makes an anomalous observation unusual. Hence, it provides support to localize the reason of the anomaly.

The proposed approach is model-based; it relies on the multivariate probability distribution associated with the observations. Since the rare events are present in the tails of the probability distributions, we use copula functions, that are able to model the fat-tailed distributions well. The presented procedure scales well; it can cope with a large number of high-dimensional samples. Furthermore, our procedure can cope with missing values, too, which occur frequently in high-dimensional data sets.

In the second part of the paper, we demonstrate the usability of the method through a case study, where we analyze a large data set consisting of the performance counters of a real mobile telecommunication network. Since such networks are complex systems, the signs of sub-optimal operation can remain hidden for a potentially long time. With the proposed procedure, many such hidden 
issues can be isolated and indicated to the network operator.
\end{abstract}

%
%



\setcopyright{acmcopyright}
\acmJournal{TIST}
\acmYear{2019} \acmVolume{1} \acmNumber{1} \acmArticle{1} \acmMonth{1} \acmPrice{15.00}

\maketitle

\section{Introduction}

Anomaly detection refers to the process of identifying  unexpected objects or patterns, which do not conform to the usual behaviour. On the other hand, there is no canonical definition of anomaly, generally an object is called anomaly if it is different from normal instances with respect to its features and it is rare in the dataset \cite{goldstein2016comparative}.  In the literature, these instances are also often referred to as outliers, rare/extreme events, discordant objects \cite{xu2018comparison}.  The detection of "not-normal" observations has attracted a lot of research interest from the machine learning community since it has a wide variety of practical applications, including network intrusion, credit card fraud, health anomaly detection and so on \cite{chandola2009anomaly}.

We can distinguish between three main types of anomaly detection setups based on the availability of labels in the dataset \cite{goldstein2016comparative}. Supervised anomaly detection means that the data is fully labelled and an ordinary classifier can be used after dealing with the unbalanced class distribution \cite{domingues2018comparative}.  For semi-supervised anomaly detection, the training data consists of normal instances without any anomalies, one-class classification algorithms can be used \cite{li2003improving}, moreover, density estimation methods can be also used to model the density function of the normal class \cite{latecki2007outlier,laxhammar2009anomaly}, then the instances that do not conform to the normal profile are identified as anomalies. Unsupervised anomaly detection is performed on unlabeled data, taking only intrinsic properties of the dataset into account \cite{goldstein2016comparative}.

Anomaly detection algorithms can be further classified based on the specific data types and domains that they are suitable for, such as time series, categorical attributes, item sets, graphs, spatial data \cite{zimek2012survey,goldstein2016comparative}. The present study focuses on unsupervised anomaly detection for numerical high-dimensional data in Euclidean space. Due to the "curse of dimensionality" and the presence of irrelevant attributes high-dimensional anomaly detection is particularly challenging \cite{zimek2012survey}. In this paper, we propose a probabilistic model-based unsupervised anomaly detection method that works well with a large number of high-dimensional numerical observations. By numerical data we mean that all variables are non-categorical and have a continuous distribution.
This paper also describes a possible way to model the joint probability density based on a general real-valued multidimensional dataset. The proposed modelling approach can be used to assign anomaly scores to any point of the multidimensional space: observations from low density areas of the space receive high anomaly scores and vice versa.

Throughout the years, several algorithms have been proposed for unsupervised anomaly detection in high-dimensional numerical data, for extensive reviews and comparative studies we refer the reader to \cite{aggarwal2015outlier, zimek2012survey, goldstein2016comparative, xu2018comparison}.

A high number of outlier detection approaches rely on the concept of neighborhood,  either using $k$-nearest neighbor-distances \cite{angiulli2002fast} or using a density based approach, i.e. comparing the number of instances in $\varepsilon$-neighborhood of the object to the $\varepsilon$-neighborhood of the object's neighbors, e.g. Local Outlier Factor (LOF) \cite{breunig2000lof}. Although various variants of LOF have been proposed, it is important to note that neighborhood is less meaningful if the dimension  is high \cite{zimek2012survey}. A possible approach to handle high-dimensionality is to use dimension reduction methods to improve outlier detection \cite{ nguyen2010feature}. The assumption of dimension reduction based outlier detection methods is that a single subspace is sufficient to identify anomalies.

Another related approach is using ensemble learning methods and combine different subspaces for anomaly detection. The idea of "feature bagging" method is to derive outliers in randomly selected feature subsets using multiple outlier detection algorithms and combine the outputs to achieve more stable and effective results by avoiding the curse of high dimensionality \cite{lazarevic2005feature,nguyen2010mining}. A sequential ensemble-based framework was proposed to mutually refine feature selection and outlier scoring~\cite{pang2018sparse}.

Subspace outlier detection techniques define outlierness with respect to specific subspaces based on the observation that outliers are often embedded in locally relevant subspaces  but in the full-dimensional space they are covered by irrelevant attributes \cite{aggarwal2013high,garule2015outliers}. The challenge is to identify the relevant subspaces. Several techniques have been proposed for subspace identification recently, including rarity based techniques \cite{zhang2014outlier}, density-based and grid-based subspace clustering approaches   \cite{muller2008outrank,kriegel2009clustering}. Subspace outlier detection (SOD) is a method that works without a previous clustering step \cite{kriegel2009outlier}.  Another approach is to search for correlated subspaces of features \cite{nguyen20134s, nguyen2013cmi}, it was also proved that the subspace search problem can be transformed into a problem of clique mining for highly correlated features. In a recent paper a new subspace analysis approach  was proposed named Agglomerative Attribute Grouping (AAG)  that relies on a novel multi-attribute information theoretical measure which evaluates the "information distance" between groups of features \cite{bacher2017information}.

Detecting outlying subspaces is also advantageous for the purpose of outlier interpretation, i.e. identifying the attributes that contribute the most to abnormality \cite{vinh2016discovering}. Other methods have been also proposed to address the issue of outlier interpretation, such as Local Outlier Detection with Interpretation (LODI) \cite{ dang2013local} and Contextual Outlier INterpretation (COIN) \cite{ liu2017contextual}.

Based on the observation that angles between data points are more meaningful in high dimension than distances, angle-based outlier detection methods have also been proposed \cite{kriegel2008angle}. Angle-based methods rely on the assumption that the variance of angles between an outlier and other data points is lower than for normal observations, since for normal data points, other objects are distributed in all direction, while for outliers, they are more concentrated in some direction. A random projection based efficient approximation variant was proposed in \cite{pham2012near}.

Probabilistic anomaly detection methods have also been extensively used for high-dimensional data. The general scheme is to estimate the density function of a dataset $X$ by calibrating the model parameters $\theta $ and identifying outliers as the observations having the smallest likelihood $\mathbb{P}(X|\theta)$. The density function can be estimated by Gaussian Mixture Model using Expectation-Maximization algorithm \cite{tarassenko1995novelty}, by Dirichlet Process Mixture Model \cite{fan2011unsupervised}, by Kernel Densisty Estimators \cite{tarassenko1995novelty}, and by Robust Kernel Density Estimator \cite{kim2012robust} among others.

Isolation methods are also quite widespread, isolation forest compute an anomaly score using random forest \cite{liu2008isolation,liu2012isolation,liu2010detecting}. Instances which are easy to isolate, i.e., have short average path lengths on the isolation trees, are considered anomalous. To reduce the dimension of the feature space, only a subset of the features is considered based on their kurtosis, since it is observed that kurtosis is sensitive to the presence of anomalies.

Neural networks are also used for anomaly detection in high-dimensional data. A reconstruction-based self-organizing neural network, called Grow When Required \cite{marsland2002self} can be applied for outlier detection \cite{domingues2018comparative}, furthermore,  Recurrent Neural Networks have also been used for network traffic anomaly detection \cite{radford2018network}. A Pattern Anomaly Value (PAV) \cite{chen2008multi} based anomaly detection method for sensor data was introduced in \cite{erdmann2018unsupervised}.

We can observe that a high number of unsupervised anomaly detection methods have been proposed throughout the years. On the other hand, it is important to note that there is no canonical method, mainly because it is really difficult to interpret and compare various methods. The continuous outlier scores provided by most of the algorithms vary widely in in their scale, range, and  meaning \cite{zimek2012survey}.  Evaluation of unsupervised anomaly detection methods is a notorious problem: efficiency, effectiveness, interpretability, scalability, memory consumption and robustness should also be considered. For some possible evaluation measures, strategies and empirical evaluation of various anomaly detection techniques on benchmarks datasets we refer to \cite{zimek2012survey,campos2016evaluation, xu2018comparison}.

In this paper, we introduce a method, which tackles the high dimensional
feature space by finding some special two dimensional subspaces, which also aims to minimize redundancy. An other
important benefit of the presented approach is that it does not only give an anomaly score but also indicates why exactly an observation is
anomalous, making it easier to find and fix the source of the problem. Furthermore, the proposed method scales well for high dimension and for large sample size, and is able to return anomaly scores for samples with missing data as well.

\section{Outline of the procedure}

Our procedure is based on probabilistic modeling. The main idea is to determine the joint distribution of the random variables corresponding to the features, and to assign anomaly scores to the observations based on the density of the observations. The subspaces where an observation deviates from the majority of the samples indicate which variables are involved in the anomaly. However, the realization of this high-level description of the algorithm faces some challenges that the proposed approach aims to overcome: 

\begin{itemize}
    \item Obtaining the joint distribution of a high-dimensional dataset is difficult. Simple solutions for this problem aim to fit the dataset by a multivariate Gauss distribution or by a mixture of multivariate Gaussians. However, Gaussian density functions have a fast decay, they fail to capture the heavy-tailed behavior where most anomalies take place, making them inappropriate to use for anomaly detection.
    \item Even if the joint density function of the dataset is available, translating the density (a number between zero and infinity) to an anomaly score (between zero and one) having a physical interpretation, is not straightforward.
    \item Providing aid to localize the reason of the anomaly in a high-dimensional space, such that a human can interpret it and take the necessary actions, is also challenging.
\end{itemize}

\begin{figure}
    \centering
    \includegraphics[width=0.90\textwidth]{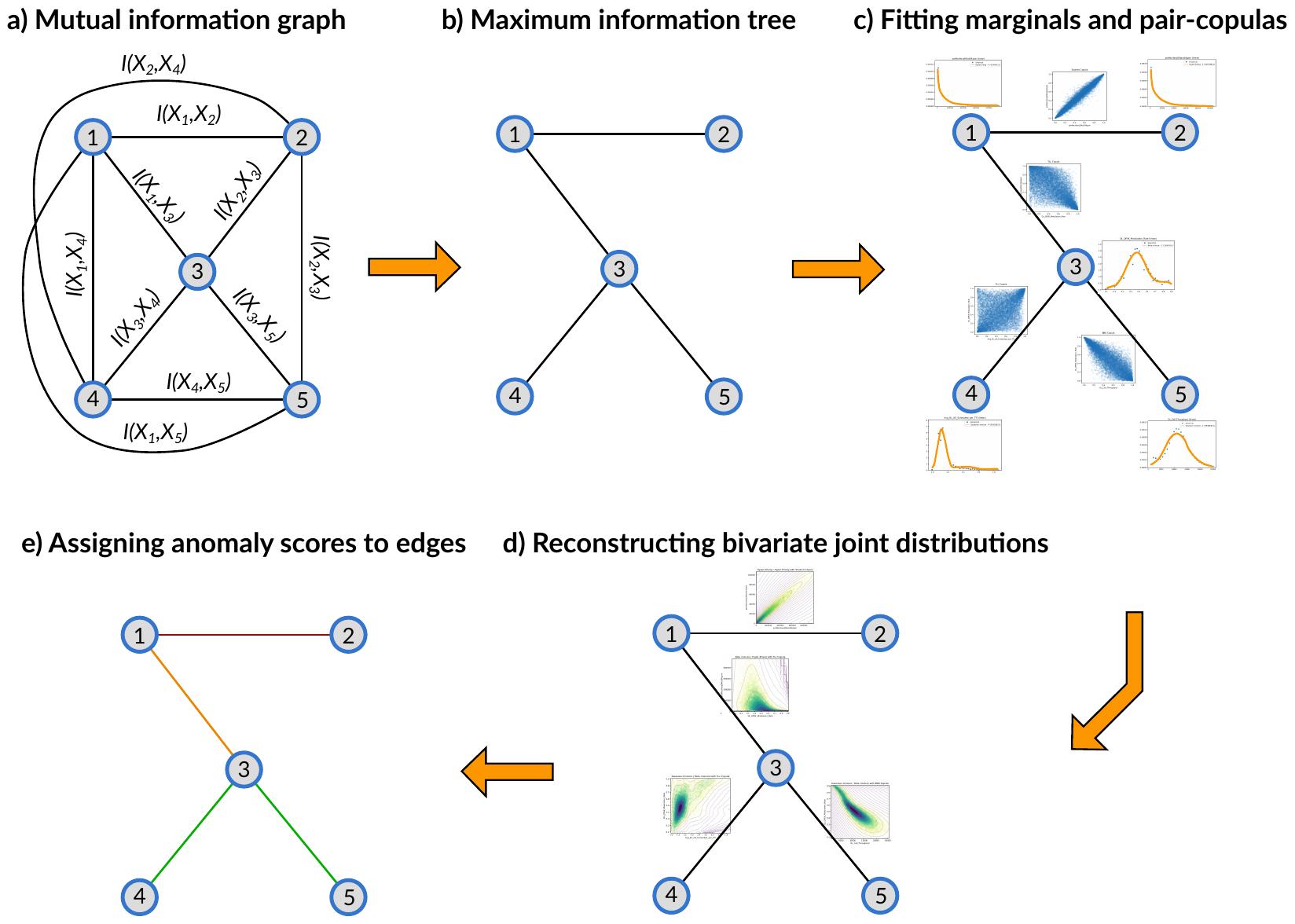}
    \caption{Concept of the presented anomaly scoring method}
    \label{fig:concept}
\end{figure}

In our approach (see Figure \ref{fig:concept}), the high-dimensional problem is decomposed into several two-dimensional ones. As the first step, we identify the most relevant \emph{pairs of variables} whose joint behavior retains as much information from the high-dimensional joint behavior as possible. The details of this step are provided in Section \ref{sec:decomposition}.

In the next step, described by Section \ref{sec:copula}, the joint distributions of the selected pairs of variables are characterized by a copula-based method.

Finally, the anomaly scores are calculated from the joint densities, as presented by Section \ref{sec:score}. For each observation, our procedure is able to report both an overall anomaly score, and individual anomaly scores for the selected variable pairs. Knowing the variable pairs affected helps to localize the reason of the anomaly.

\section{Decomposition of the high dimensional space }\label{sec:decomposition}

The direct modeling of the joint behavior of high-dimensional observations with reasonable accuracy is intractable.
To cope with the exploding complexity, we use only some of the two-dimensional marginal probability distributions in the presented method, those that are the most relevant from the anomaly detection point of view.

The question that naturally arises is how to chose a relatively small number of pairs from all possible pairs of random variables, while retaining as much information as possible. The answer is inspired by the work of Chow and Liu \cite{chowliu},  where a similar problem has been solved for discrete joint probability distributions. In this section, we provide a summary on this method, and adapt it to the case of continuous variables.

Let us consider $\mathbf{X}=( X_{1},\dots ,X_{d})^T$ a $d$-dimensional discrete random vector with probability distribution $P(\mathbf{X})$. The probability of a realization $\mathbf{x}$ is denoted by $P(\mathbf{x})$, and the set of indices of the random variables are denoted by $V=\{ 1,\dots ,d\}.$ Furthermore, we denote the probability distributions of the two-dimensional marginals by $P( X_{i},X_{j})$, $i,j\in V,i\neq j$.

In \cite{chowliu}, a product-form probability distribution has been introduced, associated with a spanning tree over the set of indices, defined by
\begin{align}
\label{eq:span_tree}
P_{Ch\_L}(\mathbf{X}) =\frac{\prod\limits_{(i,j)\in T}P(X_{i},X_{j}) }{\prod\limits_{i\in V}\left[ P( X_{i}) \right]^{v_{i}-1}},
\end{align}
where $T$ is a spanning tree defined on the vertex set $V=\{ 1,\dots,d\} $, and $v_{i}$ denotes the number of such two-dimensional marginal probability distributions in the numerator of \eqref{eq:span_tree} which
contain variable $X_{i}$. This type of probability distribution has the property that its two-dimensional marginals present in its formula coincide with the two-dimensional marginals of $P(\mathbf{X})$.

Our aim is to find a probability distribution $P_{Ch\_L}(\mathbf{X})$ of \eqref{eq:span_tree} to approximate the probability distribution $P(\mathbf{X}) = (X_{1},\dots ,X_{d})^{T}$, with the goodness of fitting quantified by the Kullback-Leibler divergence. Hence, the problem can be formulated as searching for a
probability distribution $P_{Ch\_L}(\mathbf{X})$ which minimizes the Kullback- Leibler divergence between $P(\mathbf{X}) $ and $P_{Ch\_L}(\mathbf{X})$.

To solve this problem, \cite{chowliu} has introduced an undirected complete graph defined on the set of vertices $V$. The edges of the graph are weighted by the mutual information of the two dimensional marginal
probability distributions corresponding to the two vertices connected. In that paper, it has been proven that the Kullback-Leibler divergence between $P(\mathbf{X}) $ and $P_{Ch\_L}(\mathbf{X})$ can be expressed as
\begin{align*}
KL(P_{Ch\_L}(\mathbf{X}), P(\mathbf{X})) =\sum_{i=1}^{d}H( X_{i}) -\sum_{(i,j)\in T}I(
X_{i},X_{j}) -H( \mathbf{X}),
\end{align*}
where $I(X_{i},X_{j})$ is the mutual information defined by
\begin{align}
\label{eq:mut_inf}
I(X_{i},X_{j}) = H(X_{i})+H( X_{j})
-H(X_{i},X_{j}),
\end{align}
and $H(\cdot)$ denotes the entropy of a random variable or a random vector.

It is easy to see that $KL(P_{Ch\_L}(\mathbf{X}), P(\mathbf{X}))$ is minimal when we take edges from the complete graph along the spanning tree having the maximum weight \cite{chowliu}, i.e., 
\begin{align*}
T^*= \underset{T}{\mathrm{argmax}}\left(\sum_{(i,j)\in T} I( X_{i},X_{j})\right).
\end{align*}

The maximum information tree $T^*$ can be obtained by applying Prim's or Kruskall's algorithm.

It is important to mention that the spanning tree structure encodes conditional independencies between the random variables associated to the vertices. If two nodes are not connected, the corresponding random variables are conditionally independent given any random variable on the path between them. 
Therefore, the best fitting $P_{Ch\_L}(\mathbf{X})$ exploits the conditional independencies existing in the multivariate probability distribution that reduces the redundancy between the corresponding random variables.

In many practical cases, such as in the case described in Section \ref{application}, the samples are obtained from a continuous probability distribution instead of a discrete one. In such cases the samples can be transformed to discrete, and the idea of Chow and Liu can still be applied.
For this transformation we use a partitioning such that each interval contains the same number of realizations for all random variables. This way we get $H(X_{i}) =H(X_{j})$, for all $i,j\in \{ 1,\dots, d\}, i\neq j$, and therefore we have that $I(X_{i},X_{j})$ depends only on $H(X_{i},X_{j})$ (see \eqref{eq:mut_inf}).

Since the maximum of $\sum_{i,j\in T}I(X_{i},X_{j})$ is obtained when $\sum_{i,j\in T}H(X_{i},X_{j})$ is minimal, we weight the edges of the complete graph by the entropy of the bivariate random variables connected, and construct the minimal weighted spanning tree of this graph. As a result, we obtain a set of $d-1$ variable pairs, whose bivariate marginal probability distributions capture as much information on the joint behavior as possible (Step b. in Figure \ref{fig:concept}).

There are software packages available to obtain the maximum information tree, also for the case of continuous variables \cite{schepsmeier2018package}. These packages use Kendall's tau instead of the mutual information for weighting the graph. However, for big data the computation of Kendall's tau is computationally expensive, therefore we recommend to use the mutual information on the discretized data for this purpose. The tree structure is not sensitive for the chosen metric, the obtained three structure provided by these methods was the same in all of the studied cases. 

\section{Copula-based modeling of pairs of variables}\label{sec:copula}

The proposed method relies on \emph{copulas} to model the joint distributions of the variable pairs obtained by the maximum information tree. Copula-based methods are widely used in modelling joint probability distributions (mainly in the field of financial mathematics) due to their benefits in representing heavy tails \cite{cherubini2004copula}. Since anomalies can be considered as events falling in the tail of the probability distribution, copulas are very useful in modeling anomalies.

According to Sklar's theorem, the joint distributions are uniquely decomposed to the marginal distributions and a so called copula function, that characterizes only the dependency structure between the variables, and is independent from the marginal distributions. Consequently, copulas enable the separate modeling of the dependence structure and the marginals; we are going to rely heavily on this feature in the presented method.

\subsection{A short overview on copulas}

Before detailing the exact kind of copulas used in this paper, let us review some concepts related to copulas (for more details see \cite{nelson2007extremes}).
\begin{definition}
A function $C:\left[ 0;1\right] ^{d}\rightarrow \left[ 0;1%
\right] $ is called a $d$-dimensional copula if it satisfies the following
conditions:
\begin{enumerate}
\item $C\left(u_{1},\dots, u_{d}\right) $ is monotonously increasing in each component $u_{i}$,
\item $C\left(u_{1},\dots, u_{i-1},0,u_{i+1},\dots, u_{d}\right) = 0$ for
all $u_{k}\in \left[0;1\right] $,\ $k\neq i,\ i=1,\dots ,d$,
\item $C\left(1,\dots, 1,u_{i},1,\dots ,1\right) =u_{i}$ for all $%
u_{i}\in \left[0;1\right] ,\ i=1,\dots ,d$,
\item $C$ is $d$-increasing, i.e for all $\left(u_{1,1},\dots, u_{1,d}\right)$ and $\left(u_{2,1},\dots, u_{2,d}\right)$ in $\left[0;1\right]^{d}$ with $u_{1,i}<u_{2,i}$ for all i, we have
\begin{align*}
\sum\limits_{i_{1}=1}^{2}\cdots\sum\limits_{i_{d}=1}^{2}\left( -1\right)^{\sum\limits_{j=1}^{d}i_{j}}C\left(u_{i_{1},1},\dots ,u_{i_{d},d}\right) \geq 0.
\end{align*}
\end{enumerate}
\end{definition}
Thus, copulas are $d$ dimensional distribution functions such that each of their marginals are uniformly distributed.

Due to Sklar's theorem \cite{schweizer1969measures}, if $X_{1},\dots ,X_{d}$ are continuous random variables defined on a common probability space, with univariate marginal cdf's $F_{X_{i}}\left(x_{i}\right) $ and joint cdf $F_{X_{1},\dots, X_{d}}\left(x_{1},\dots, x_{d}\right)$, there exists
a unique copula function $C_{X_{1},\dots, X_{d}}\left(u_{1},\dots, u_{d}\right) :\left[0;1\right]^{d}\rightarrow \left[0;1\right]$, such that by the substitution $u_{i}=F_{i}\left(x_{i}\right) ,\ i=1,\dots, d$, we get
\begin{align}
F_{X_{1},\dots ,X_{d}}\left( x_{1},\dots ,x_{d}\right) =C_{X_{1},\dots
,X_{d}}\left( F_{1}\left( x_{1}\right) ,\dots ,F_{d}\left( x_{d}\right)
\right), ~~~\text{for all $\left( x_{1},\dots ,x_{d}\right) ^{T}\in R^{d}$.}\label{eq:sklar}
\end{align}
This relation is of principal importance. It states that \emph{every} multivariate joint distribution can be fully reconstructed based on its marginals and its copula. 

The density function $f_{x_{1},\ldots ,x_{d}}(x_{1},\ldots ,x_{d})$ can be
expressed in the following way:

\begin{align}
\label{eq:wellknown}
\begin{split}
f_{x_{1},\ldots ,x_{d}}\left( x_{1},\ldots ,x_{d}\right)  &=\frac{\partial
^{d}F(x_{1},\ldots ,x_{d})}{\partial x_{1}\ldots \partial x_{d}}=\left.
\frac{\partial ^{d}C(u_{1},\ldots ,u_{d})}{\partial u_{1}\ldots \partial
u_{d}}\right\vert_{u_{i}=F\left( x_{i}\right) }\prod_{i=1}^{d}\frac{\partial F\left(
x_{i}\right) }{\partial x_{i}} \\ &=c(F_{1}\left( x_{1}\right) ,\ldots
,F_{d}\left( x_{d}\right) )\prod_{i=1}^{d}f(x_{i}),
\end{split}
\end{align}

where $c(u_{1},\ldots ,u_{d})$ is the copula density function.


Modeling high-dimensional probability distributions by using only a single high-dimensional copula to characterize all the underlying dependencies is possible, although rather restrictive, as there is a high chance that none of the currently known copula families is flexible enough to approximate the complex dependency structure of a real dataset with reasonable accuracy (as shown in \cite{aas2009pair}).
A new approach has been introduced by Joe \cite{joe1997multivariate}, the so called \emph{pair-copula construction}, that makes possible to capture different kinds of dependencies between the pairs of the variables in the same multivariate probability distribution. In this approach, a copula is expressed as a product of different types of bivariate copulas and bivariate conditional copulas. A useful modeling tool, following the same direction, is the so called R-vine structure, that was introduced by Bedford and Cooke \cite{bedford2001probability, bedford2002vines} and described in more details by Kurowicka and Cooke \cite{kurowicka2006uncertainty}. 
The drawback of using R-vine copulas in higher dimensions is that the number of parameters grows very fast with the dimensions. To address this issue, a special kind of vine-copula was proposed, called truncated
vine-copula. This concept has been introduced by Brechmann \cite{brechmann2010truncated} and a more general version has been presented in \cite{kovacs2016hypergraphs, kovacs2017connection}.

We do not review the R-vine graph structures here, as we need only a special case of them, the truncated vines \cite{brechmann2010truncated, brechmann2012truncated}. In our approach, we will use the so called truncated vine copula truncated at level one, that is directly related to Chow Liu probability distributions (see Section \ref{sec:decomposition}).
Let us denote the bivariate probability distribution function of $X_{i}$, $X_{j}$ by $F_{i,j}$, the bivariate probability density function by $f_{i,j}$, and the bivariate copula density function corresponding to $f_{i,j}$ by $c_{i,j}$, for $i,j\in V$.

Let us consider the approximation of the joint probability density function $f_{X_{1},\ldots
X_{d}}\left( \mathbf{x}\right) $ given by the by a Chow-Liu type density
function:
\begin{align}
\begin{split}
f_{Ch\_L}\left( \mathbf{x}\right) =\prod\limits_{e\in T, s \in V}\frac{%
f_{j(e)k(e)}\left( x_{j(e)},x_{k(e)}\right) }{f_{s}\left( x_{s}\right)
^{v_{s}-1}},
\end{split}
\end{align}
where $v_{s}$=the number of edges connected to vertex $s$ (see \eqref{eq:span_tree} for the discrete case).

The probability density function $f_{Ch\_L}\left( \mathbf{x}\right) $ can be expressed as follows by
dividing and multiplying with the marginal probability densities of the
bivariate densities in the numerator, and then by applying \eqref{eq:wellknown} to the bivariate case:

\begin{align}
\label{eq:chowliu}
\begin{split}
f_{Ch\_L}\left( \mathbf{x}\right)  &=\prod\limits_{i=1}^{d}f\left(
x_{i}\right) \prod\limits_{e\in T}\frac{f_{j(e)k(e)}\left(
x_{j(e)},x_{k(e)}\right) }{f_{_{j(e)}}\left( x_{j(e)}\right) \cdot
f_{_{k(e)}}\left( x_{k(e)}\right) }= \\
&=\prod\limits_{i=1}^{d}f\left( x_{i}\right) \prod\limits_{e\in
T}c_{j(e)k(e)}\left(F_{j(e)}\left( x_{j(e)}\right) ,F_{k(e)}\left(
x_{k(e)}\right) \right).
\end{split}
\end{align}

It is easy to see that from \eqref{eq:wellknown} and from \eqref{eq:chowliu} it
follows that $\prod\limits_{e\in T}c_{j(e)k(e)}\left(F_{j(e)}\left(
x_{j(e)}\right) ,F_{k(e)}\left( x_{k(e)}\right) \right)$ is a valid copula density.

Observe that in \eqref{eq:chowliu} only pair-copulas are involved, this is the reason why these constructions are also called pair-copula constructions. Formula \eqref{eq:chowliu} also shows that the univariate marginal probability distributions and the bivariate copulas can be fitted separately.

Using the elements introduced above, the inference of the multivariate density function is boiled down to the following steps (see also Figure \ref{fig:concept}):
\begin{itemize}
\item The selection of a specific tree-structure. For this purpose we use the maximum information tree described in Section \ref{sec:decomposition}.
\item The choice of pair-copulas involved in formula \eqref{eq:chowliu}. For each edge in the maximum information tree, thus for each variable pair, a pair-copula is constructed and its parameters are estimated.
\item The modeling of the univariate marginals. For every vertex of the maximum information tree, thus for every variable, the marginal distribution is estimated with an appropriately chosen distribution family.
\item Construction of the two-dimensional joint distributions. For every edge of the maximum information tree, the joint distribution of the variable pair is reconstructed from the two related marginals and the copula.
\end{itemize}

Let us consider i.i.d observations $D=\left\{ \mathbf{x}^{1},\ldots ,\mathbf{x}%
^{N}\right\} ,\mathbf{x}^{m}\in \mathbb{R}^{d}$. Now we fit a
Chow-Liu density function to the data. The expression of the loglikelihood function of Chow-Liu density function \eqref{eq:chowliu} fitted to the data $D$ is the following:

\begin{align}
\label{eq:loglik}
\begin{split}
\ln L\left( f_{Ch_L}\left( D\right) ,\mathbf{\alpha }_{lm}\mathbf{,\beta }%
\right)  &=\sum\limits_{e\in T}\sum\limits_{m=1}^{N}\ln
c_{j(e)k(e)}\left(F_{j(e)}\left( x_{j(e)}^{m}\right) ,F_{k(e)}\left(
x_{k(e}^{m})\right) ,\mathbf{\alpha }_{e}\right)+ \\
&+\sum\limits_{m=1}^{N}\ln f\left( x_{i}^{m},\mathbf{\beta }\right) ,
\end{split}
\end{align}
where $\mathbf{\alpha }_{e}\mathbf{,\beta }$ are the parameters of the pair
copulas in the edges, and the marginal density functions respectively. From  \eqref{eq:loglik} it follows that the loglikelihood of the Chow-Liu
approximation can be expressed as the sum of loglikelihoods of the
pair-copulas and the sum of the loglikelihood of the marginal pdf's. This makes our fitting tractable, paralelizable and flexible to multiple
kind of dependencies.

The following three subsections provide solutions for the last three items, for the fitting of pair-copulas, the univariate marginals, and the construction of the joint distribution.

\subsection{Fitting the pair-copulas}\label{sec:fitcopula}

For every variable pair in the maximum information tree, our procedure first extracts the empirical copula. To do so, both variables need to be transformed to pseudo-observations, hence they are replaced by the normalized ranked data.
The pseudo-observations corresponding to $N$ realizations of a $d$-dimensional random vector $x_{i}=(x_{i1}, \dots, x_{id})^T$, $i \in \{1, \dots, N\}$ are defined by $u_{ij}=\frac{r_{ij}}{N+1}$, $i \in \{1, \dots, N\}$, $j \in \{1, \dots, d\}$, , where $r_{ij}$ denotes the rank of $x_{ij}$ among all  $x_{kj}$, $k \in \{1, \dots, N\}$~\cite{schepsmeier2018package}. 
This way we get a two-dimensional data series where both variables are uniformly distributed individually, and their joint values are the realizations of their copula distribution.

To find the appropriate copula model for the empirical copula, we consider a set of copula families, perform fitting the observations with all of them, and select the one providing the highest likelihood.

In our method the following copula families (and their rotations) are considered:
\begin{itemize}
    \item Archimedean copulas, including the Gumbel, Clayton, Joe, Frank, BB1, BB6, BB7, BB8 copulas. These copulas have a low number of parameters and an explicit copula density, that makes the fitting process relatively efficient.
    \item Elliptical copulas \cite{frahm2003elliptical}, including the Gaussian and the Student-t copula. Contrary to the Archimedean copulas the elliptical ones are symmetric. They do not have an explicit density, hence their fitting does not scale very well with the amount of training data.
    \item Non-parametric copula: the TTL copula. It makes use of a local-likelihood transformation kernel density estimation (see \cite{geenens2017probit}). This method has a large number of parameters (900) and yields a less smooth density; in the presented procedure it serves as a remedy when the empirical copula is so asymmetric that the above listed copula families fail to fit it.
\end{itemize}

For this step of the procedure we were relying on the built-in copula fitting capabilities of the vinecopulib package\footnote{\url{https://github.com/vinecopulib/vinecopulib}}.

\subsection{Fitting the univariate marginal distributions}\label{sec:fitmarginal}

The concept of fitting the marginals for each variable (a vertex in the maximum information tree) is the same as in case of copulas: having defined a wide set of distribution families we repeat the fitting with all of them and choose the one that has the highest likelihood for the training data.

The following univariate distribution families were the candidates for fitting the marginals:
\begin{itemize}
    \item Simple univariate distributions like the exponential, student-t, log-normal, and the Pareto distributions. Finding the maximum likelihood estimation for these distributions is simple, but they often do not have the flexibility to fit real-word dataset accurately enough.
    \item Gaussian mixture models. Gaussian mixture models are widely used for fitting real datasets. With the  expectation-maximization algorithm it is possible to find its optimal parameters rather efficiently. However, the density function of Gaussian mixtures has a fast decay, they are not the optimal choice for distributions having a heavier tail.
    \item Hyper-Erlang distributions. Hyper-Erlang distributions are the mixtures of Erlang distributions. Efficient expectation-maximization algorithm for the inference is available for these distributions, too (see \cite{thummler2005novel}). While the gamma distributions are more general than the Erlang distributions, interestingly we got far better fitting results with hyper-Erlangs. In fact, they turned out to be the most flexible choices for the majority of the variables in our use case in Section \ref{application}.
    \item Mixture of beta distributions. To model distributions with finite support, the mixture of beta distributions provide an appealing option. We have successfully applied the hybrid EM -- moment fitting method published in \cite{schroder2017hybrid}.
\end{itemize}

\subsection{Construction of the joint density functions}\label{sec:fitjoint}

Once the univariate marginals and the pair-copulas are available, the two-dimensional joint densitiy functions can be reconstructed based on \eqref{eq:wellknown}, yielding
\begin{align}
    f_{X_i,X_j}(x_i,x_j) = c_{X_i,X_j}(F_{X_i}(x_i), F_{X_j}(x_j))\cdot f_{X_i}(x_i) \cdot f_{X_j}(x_j),
\end{align}
and the overall density of a $d$ dimensional observation is given in product form by \eqref{eq:chowliu}.




\section{Anomaly scores derived from the joint probability distribution}\label{sec:score}

Let us consider a $d$-dimensional random vector $X=(X_{1},\dots X_{d})^{T}$ taking values in $\chi \subset\mathbb{R}^{d}$, with a bounded probability density function denoted by $f(\mathbf{x})$. The aim of this section is to define an anomaly scoring function which is defined on the $d$-dimensional support of the random vector $\mathbf{X}$ and takes values in the $[0;1]$ interval. 

There are several distributions for which it is easy to obtain an anomaly score from the density functions. In the case of a symmetrical probability distribution, such as the univariate Gaussian distribution, it is known that the realizations which deviate from the mean value more than $2\sigma $ are considered rare. The probability that a realization of a Gaussian random variable takes place in the set $A_{0.9545}^{G}=\{x\ |\ |x-\mu|<2\sigma \}$ is $0.9545$. For non-symmetrical, uni-modal distributions, a plausible set $A_{0.9545}$ would contain values between the $2.275$ and $97.725$ percentiles. 

For multivariate distributions with multimodal density functions, however, it is not obvious how to define such an anomaly scoring function. Results related to this topic were developed in \cite{einmahl1992generalized, polonik1995measuring}, based on the idea of
splitting the feature space in two halves, and observations not falling into \emph{the minimum volume set} (see Definition 5 in \cite{garcia2003level}) are considered to be anomalous. In our approach we do not only classify the realizations as normal or anomalous but we also assign anomaly scores to them. Other recent work in this direction can be found in \cite{clemenccon2017mass}, where a so called \emph{Mass Volume Curve} (MV curve) is defined. The idea behind MV curves is to define an ordering on $\chi \subset \mathbb{R}^{d}$ according to which realizations of $\mathbf{X}$ can be ranked related to how rare they are.

In this paper we rely on the \emph{Mass Volume Curve}, too. In this section we review the main elements of this approach and provide and algorithm to obtain the anomaly scores of any high-dimensional observations based on the joint density function efficiently.

Let us start with the definition of the scoring function and the level set.
\begin{definition}[from \cite{clemenccon2017mass}]
 A \emph{scoring function} is any measurable function $s:\chi\rightarrow\mathbb{R}_{+}$ that is integrable with respect to the Lesbesque measure.
 
A \emph{level set} corresponding to a scoring function $s$ and a level $t$ is given by $A_{t}=\{\mathbf{x\in \chi \ |\ }s(\mathbf{x})\geq t\}, t\in [0,\infty]$ \footnote{In \cite{clemenccon2017mass} $t\in[-\infty ,\infty]$}.
\end{definition}
Note that the "scoring function" as defined above is not the same as our anomaly scoring function defined later. It is easy to see that if $t\geq t'$ then $A_{t}\subset A_{t'}$. Next, we review the two most important notions, the \emph{mass} and the \emph{volume} of a level set.

\begin{definition}
For a scoring function $s$ and a level $t$, $\alpha _{s}\left( t\right)
=P\left( s\left( \mathbf{X}\right) \geq t\right) $ is called the mass of the level set $A_{t}$.
\end{definition}

\begin{definition}
For a scoring function $s$ and a level $t$, $\lambda _{s}\left( t\right)
=\lambda \left( \left\{ \mathbf{x}\in \chi \ |\ s\left( \mathbf{x}\right)
\geq t\right\} \right) $ is called the volume corresponding to a level $t$, with respect to the Lesbesque measure.
\end{definition}

Based on \cite{polonik1995measuring}, we consider the density function of $\mathbf{X}$ as a scoring function. In this case $\lambda_{f}(t) $ is the contour cluster of the density function $f(x)$ at level $t$.

\begin{figure}
    \centering
    \includegraphics[scale=0.6]{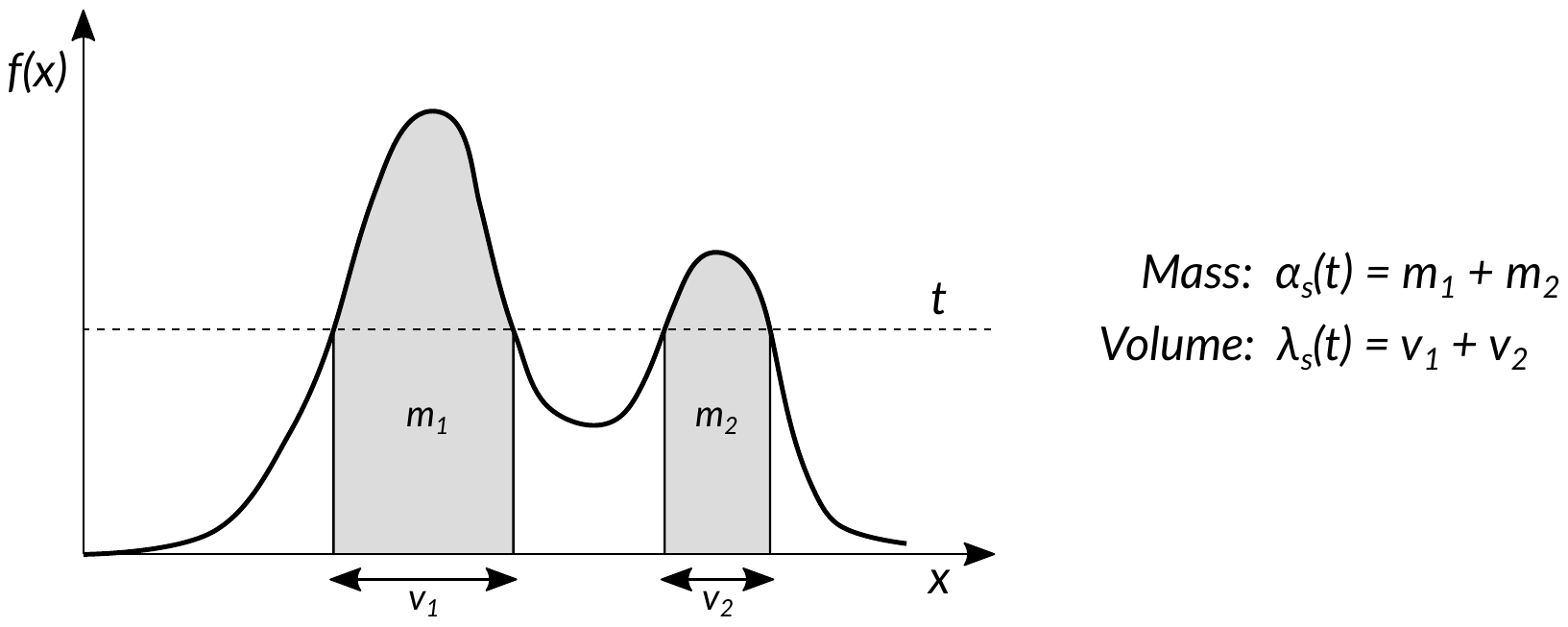}
    \caption{Mass and volume on a bi-modal density function}
    \label{fig:mvbimod}
\end{figure}

Mass, volume, and their relation with anomaly scoring can be interpreted easier though an example. In Figure \ref{fig:mvbimod} there is a bi-modal univariate density function, and a level $t$. The mass corresponding to level $t$ is the integral of the density function where it is greater than level $t$, and the volume covers the intervals on the $x$ axis where the density function is above level $t$. Hence, the volume covers the most dense part of the density function and the mass is the probability that a realization falls into this part.

The problem of anomaly detection can be formulated via the so called minimum volume sets. Minimum volume set is a concept introduced in \cite{einmahl1992generalized, polonik1995measuring} as a solution of the following constrained minimization problem
\begin{align}
A_\alpha^* = \underset{A\subset\chi}{\arg\min}~\lambda(A), ~~~\text{such that $P(\mathbf{X}\in A)\geq\alpha$},\label{eq:aalphas}
\end{align}
thus, we are looking for the smallest possible volume yielding mass $\alpha$. In \cite{clemenccon2017mass} it is stated that if the probability density function is bounded and has no flat parts, for any $\alpha \in \left] 0;1\right[ $ there exists a unique minimum volume set $A_\alpha^\ast$ whose mass is equal to $\alpha$. The "rare realizations" are those that belong to the complementary set, i.e.,  $\chi \backslash A_\alpha^*$, when $\alpha$ is sufficiently large (close to $1$).

After these preliminaries we can define the theoretical anomaly score as follows.
\begin{definition}
The theoretical anomaly score of a realization $\mathbf{x}$ is
defined by
\begin{align}
\mathcal{A}(\mathbf{x}) = \inf\{\alpha: \mathbf{x}\in A_\alpha^*\}.
\end{align}
\end{definition}

It is easy to see that $\mathcal{A}\left( \mathbf{x}\right) \in \left[ 0;1\right] $, and the larger the value of $\mathcal{A}\left( \mathbf{x}\right) $ is the more rare the realization $\mathbf{x}$ is.
The main problem is, however, the numerical computation of $A_\alpha^*$ involving the solution of \eqref{eq:aalphas}.

To overcome this problem we will now define an approximate anomaly score. To do so we generate i.i.d. random samples from $f(\mathbf{x})$, denoted by $\tilde{\mathbf{x}}^{i},\ i=1,\dots, m$, where $m$ is the sample size.
To each vector $\tilde{\mathbf{x}}^{i}$ we assign its density $f(\tilde{\mathbf{x}}^{i})$. This way we get a univariate
discrete random variable $\tau$, which takes values $f(\tilde{\mathbf{x}}^{i})$ with probability $p_{i}=\frac{1}{m}$.
Then, the anomaly score is approximated by the complementary distribution function of this random variable (denoted by $\bar{G}(t)=P(\tau>t)$) at $t=f(\tilde{\mathbf{x}}^{i})$ .
\begin{definition}
The approximated anomaly score of any realization $\mathbf{x}$ is defined by
\begin{align}
    \hat{\mathcal{A}}(\mathbf{x}) = \bar{G}(f(\mathbf{x})) = \frac{1}{m}\sum 1_{\{\tau > f(\mathbf{x})\}}.
\end{align}
\end{definition}

\begin{figure*}
    \includegraphics[width=0.345\linewidth]{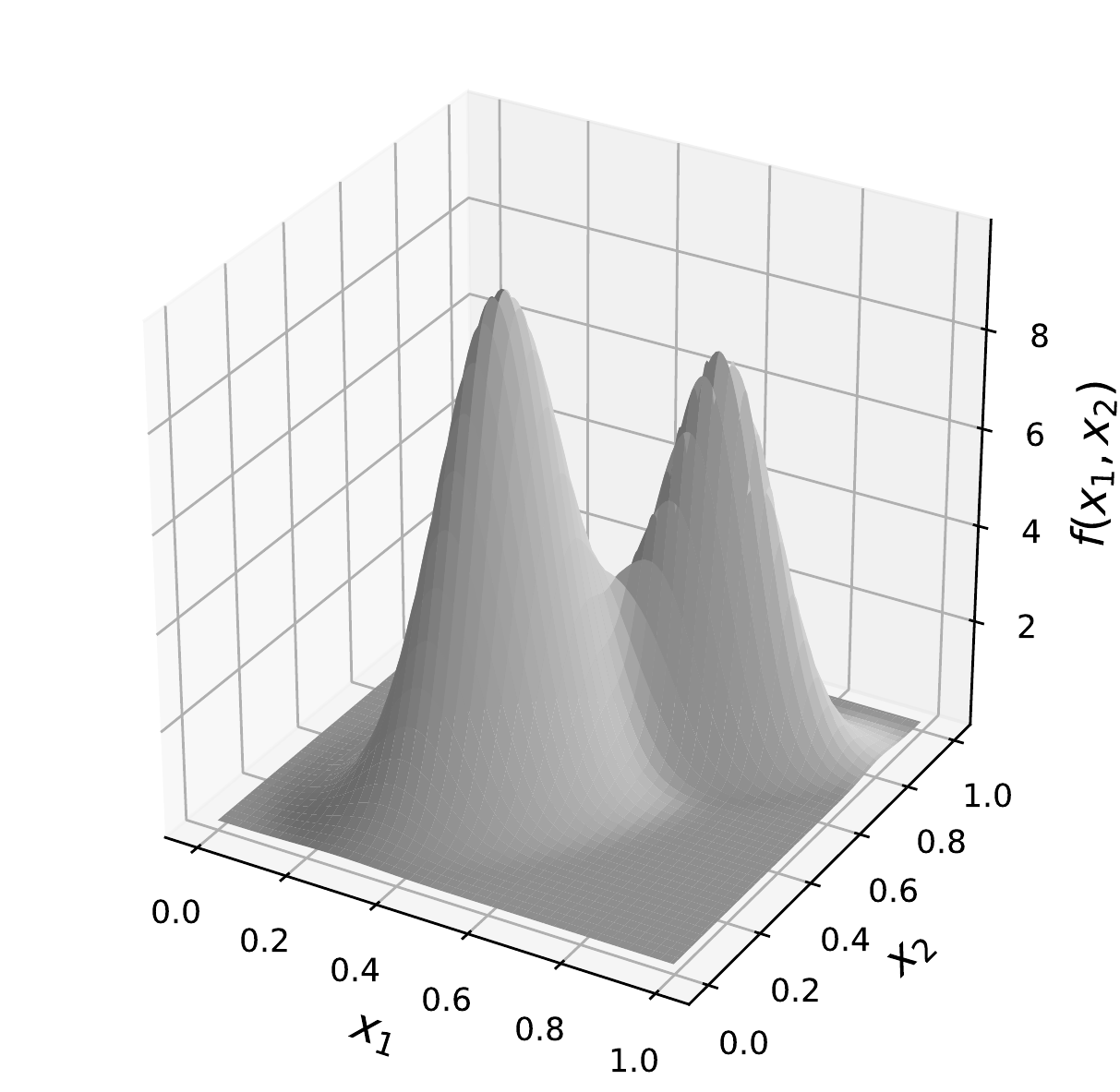}\hfill
    \includegraphics[width=0.29\linewidth]{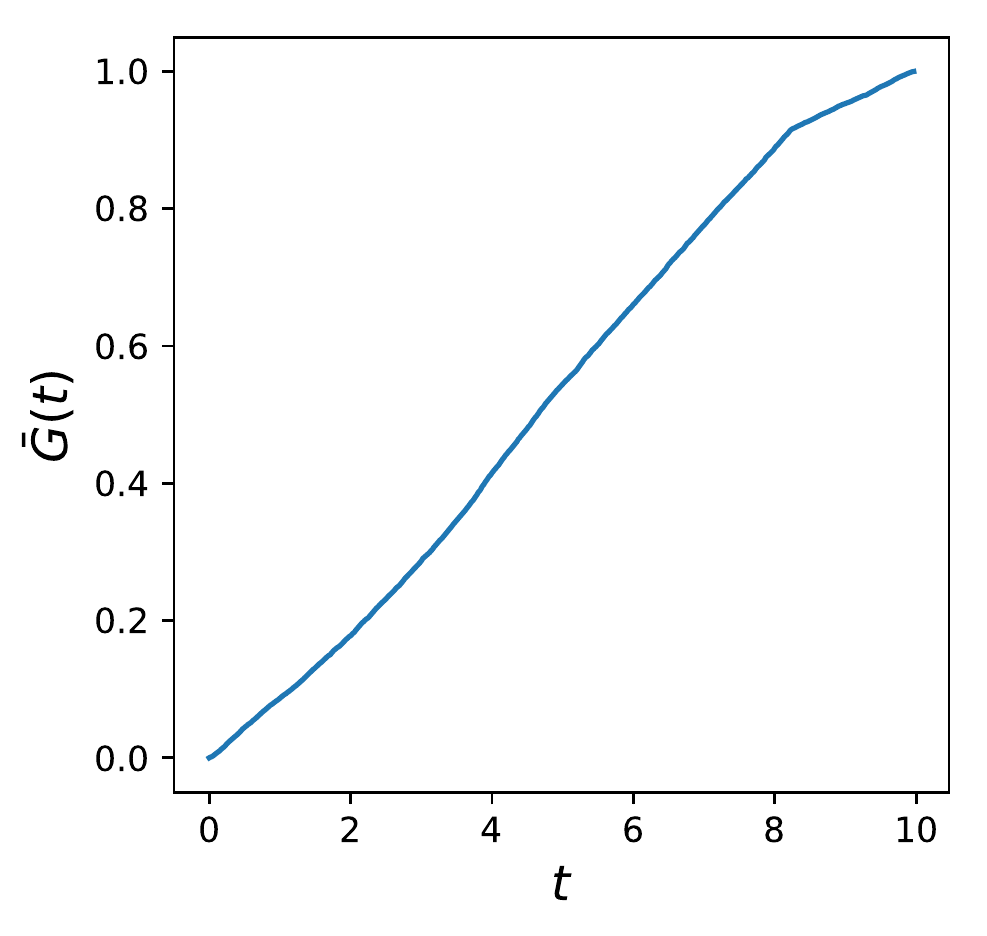}\hfill
    \includegraphics[width=0.345\linewidth]{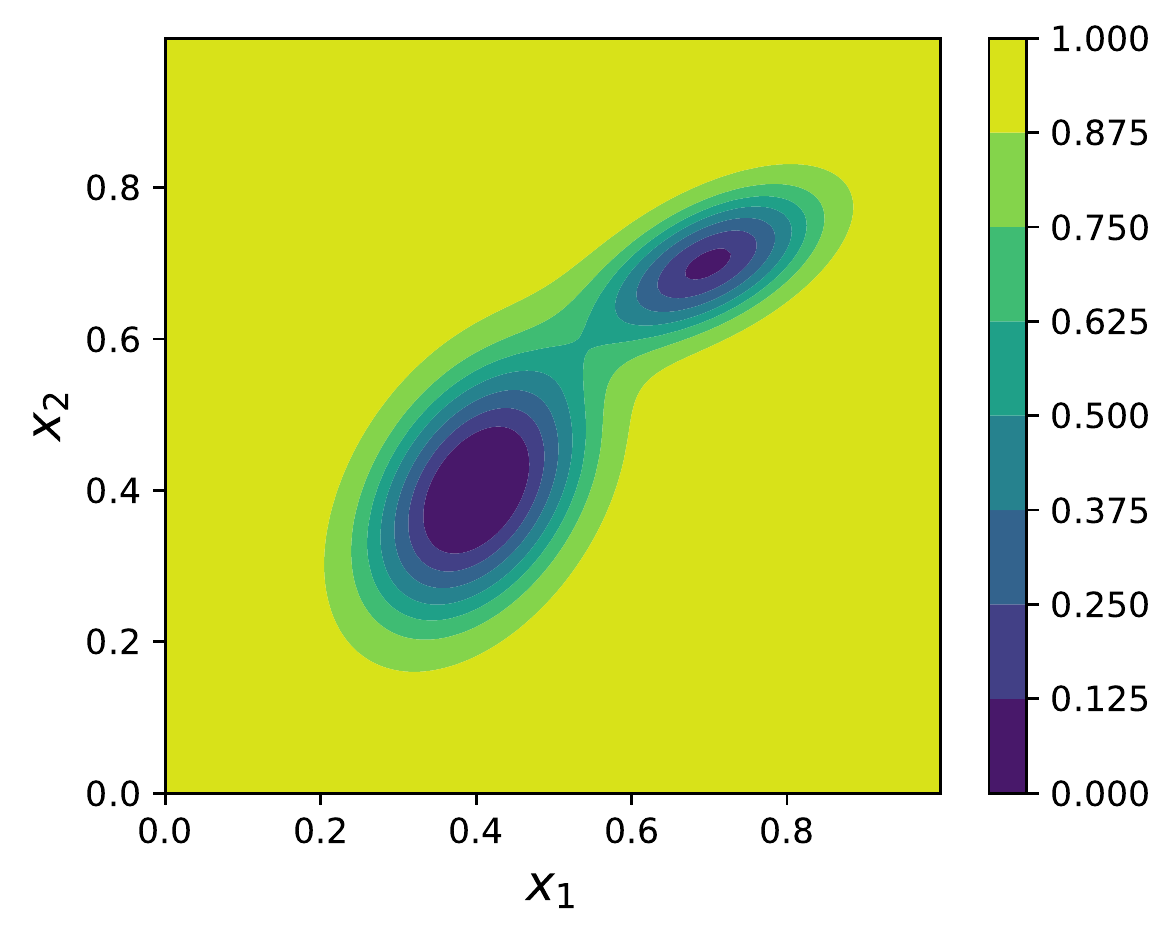}
    \caption{Obtaining anomaly score based on the pdf, in case of a 2-dimensional Gaussian mixture distribution. Left-most: the density, in the middle: $\bar{G}(t)$, right-most: the anomaly scores of the two-dimensional realizations}
    \label{fig:asgsm}
\end{figure*}

It is clear that the approximated anomaly score takes values in $%
\left[ 0;1\right]$, and has the property to be large for the rare events. 
An example for the anomaly score corresponding to a two-dimensional Gaussian mixture is illustrated in Figure \ref{fig:asgsm}.

This procedure can be used to identify anomalous cases based on some of the bivariate projections of the multivariate pdf, and also enables the localization of the anomaly, by highlighting those variables whose relation is unusual.

Whenever we have an incomplete realizations, i.e. the observed vector contains missing values, some links of the tree are missing, but all the others can be interpreted as in the case without the missing values. Hence, we can still obtain anomaly scores based on the unusual relation of the existing values.


\section{Application of the new anomaly scoring method to real mobile network data}
\label{application}

In this section we illustrate the applicability of the proposed procedure on a real dataset originating from the live commercial LTE network of a Western European mobile operator.
Mobile networks are complex systems with many parameters, where discovering that the system performance is sub-optimal is difficult for a human operator.

For decades, telecommunication operators have relied on network domain experts to report problems that affect the network performance and customer experience, with the aim of performance monitoring tools. The conventional approach of the tools is to pre-select a set of Key Performance Indicators (KPIs), based on human knowledge and experience of the domain experts. The trending of these KPIs are closely monitored, based on pre-defined single-value or multi-value thresholds, and/or pre-built single variate or multivariate time series profiles. If the KPI values exceed the thresholds/profiles, then alarms will be raised to trigger investigation, mitigation and on-site support. The experts also try to derive Root Causes Analysis (RCA) on these problems to fix as early as they can. All the processes are manual or semi-auto with limited support of tools. The traditional tools are mainly based on human knowledge, and that the complex and hidden rules in the telecommunication network are not easy to fully detect, capture and utilize.

Thus there is a strong need to revolutionize network management with AI/machine learning technology. This is essential as today's communication networks are extremely complex systems consisting of hundreds of thousands network elements organized in cooperating, coexisting and overlapping technology layers. The network elements generate huge amounts of versatile data for performance monitoring, optimization and troubleshooting purposes. It is already quite cumbersome to tackle these tasks with traditional approaches and the effects will be even more emphasized in case of 5G. Automated solutions are needed that are capable to analyze the raw data and draw conclusions, generate actionable insights using AI technology. One important area in this field is the predictive detection of anomaly patterns that appear in the data.

Our dataset comprises a large number of performance counters collected at thousands of base stations, recorded in every hour. The original set of counters has been filtered based on data quality requirements, which resulted in $54$ counters and $1,500,000$ observations. These counters represent a wide variety of measurements: related to the PDCP (Packet Data Convergence Protocol), RLC (Radio Link Control), MAC (Media Access Control) and PHY (Physical) layers, to the handovers, the activity of the schedulers, the UL (upload) and DL (download) throughput , delay, UL interference etc.
As mentioned earlier in the paper, anomaly detection for such high dimensional data is a challenging problem. In the following subsections we present how our new method performs with this mobile network dataset.

\subsection{Selecting the relevant KPI pairs}

\begin{figure}
    \hspace{-1cm}
    \includegraphics[width=1.05\linewidth]{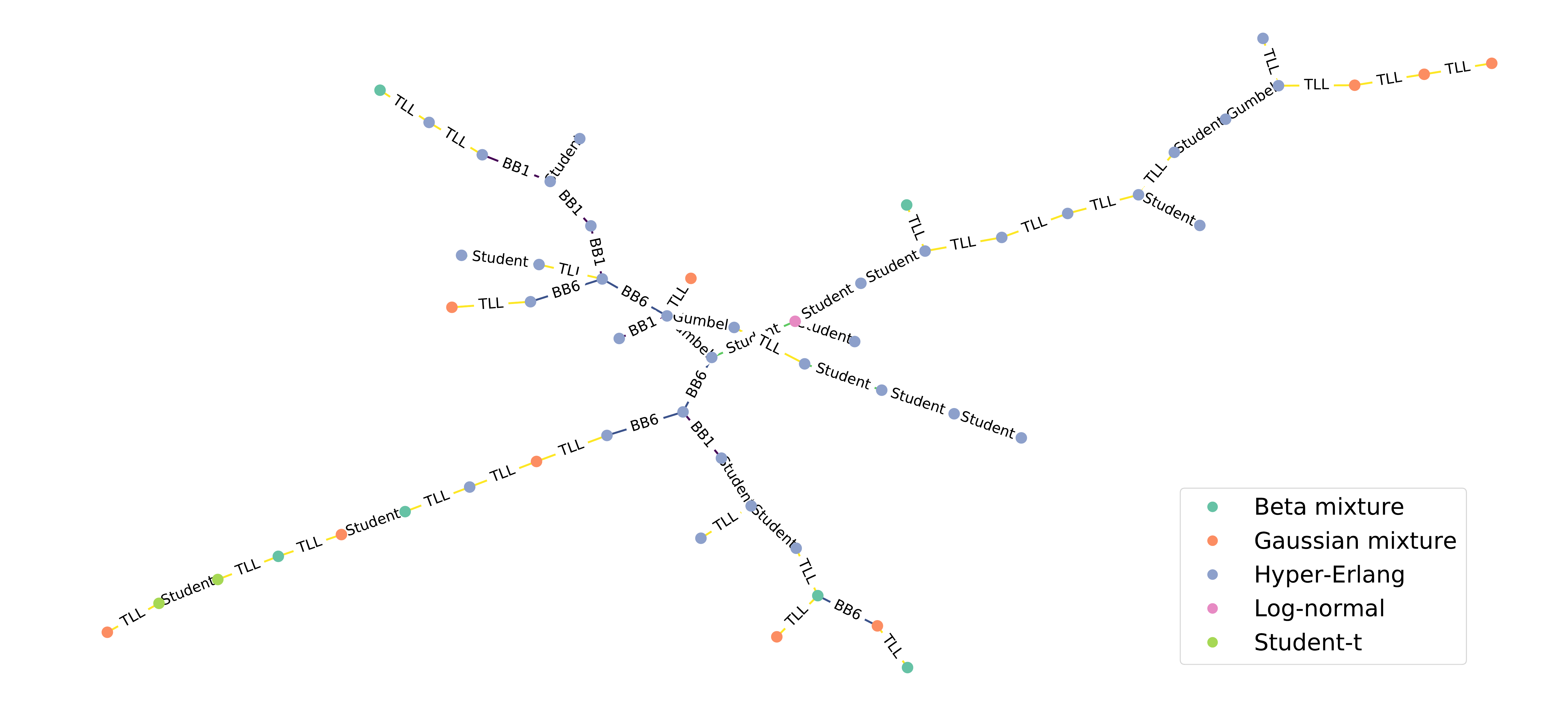}
    \caption{Maximum information tree of the KPI pairs}
    \label{fig:spanning}
\end{figure}

According to Section \ref{sec:decomposition}, the anomaly analysis of the high-dimensional problem is decomposed to many computationally tractable two-dimensional problems which can be treated separately. The first step of the procedure is the selection of the appropriate two-dimensional KPI pairs. For this purpose we create a graph where the vertices are the KPIs and the edges are weighted by the mutual information between them. Those KPI pairs are selected, which are lying on the maximum weight spanning tree of the graph. For the particular dataset considered in this example the maximum weighted spanning tree (referred to as the maximum information tree) is depicted in Figure \ref{fig:spanning}.

The two-dimensional behavior of the selected KPI pairs is then modeled by two-dimensional distributions, using bivariate copula families. The copula based approach enables the separation of the fitting problem of the marginal distributions and the fitting problem of the dependence structure between them.

\subsection{Fitting distributions to univariate marginals}

As described in Section \ref{sec:fitmarginal}, we defined various distribution families, and fitted the data of the various KPIs by maximizing the likelihood. Among the candidate distributions the one having the highest likelihood has been selected. From the total number of $54$ KPIs the log-normal distribution gave the best likelihood in $1$ case, the student distribution in $2$ cases, the beta mixture in $6$ cases. The hyper-Erlang and the Gaussian mixture distributions turned out to be the most versatile, providing the best fit in $35$ and $10$ cases, respectively. The node color in Figure \ref{fig:spanning} reflects the distribution family that fits the corresponding KPI the best.

\begin{figure}[h]
    \centering
    \includegraphics[width=0.98\linewidth]{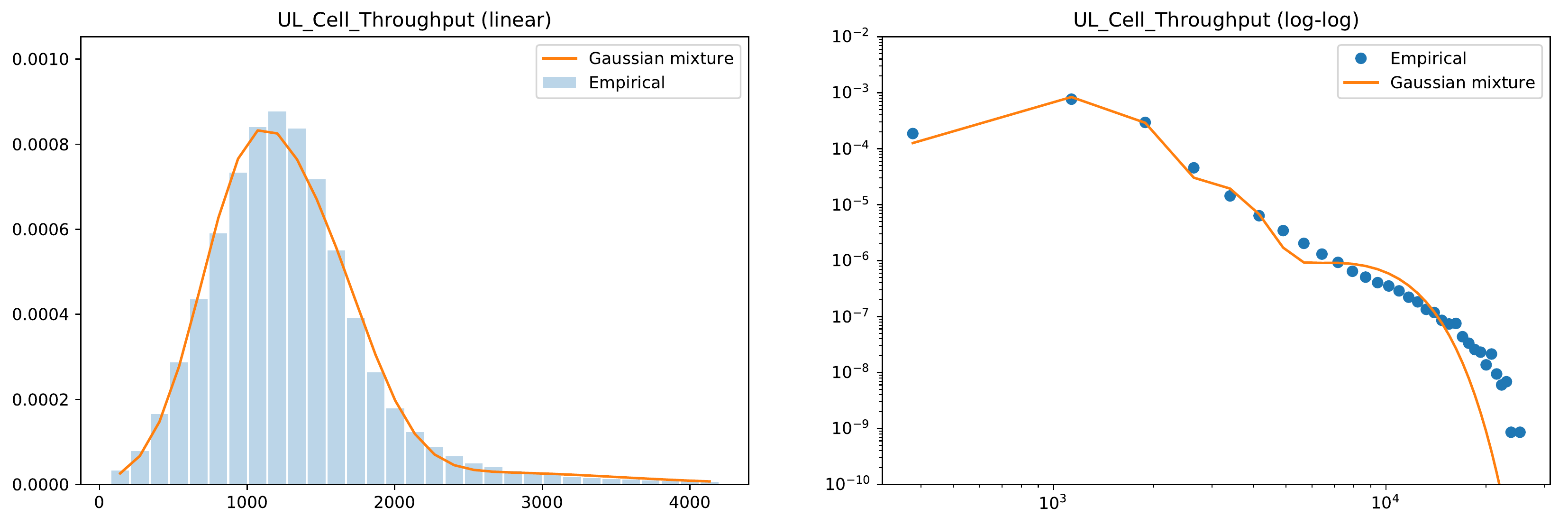}
    \includegraphics[width=0.98\linewidth]{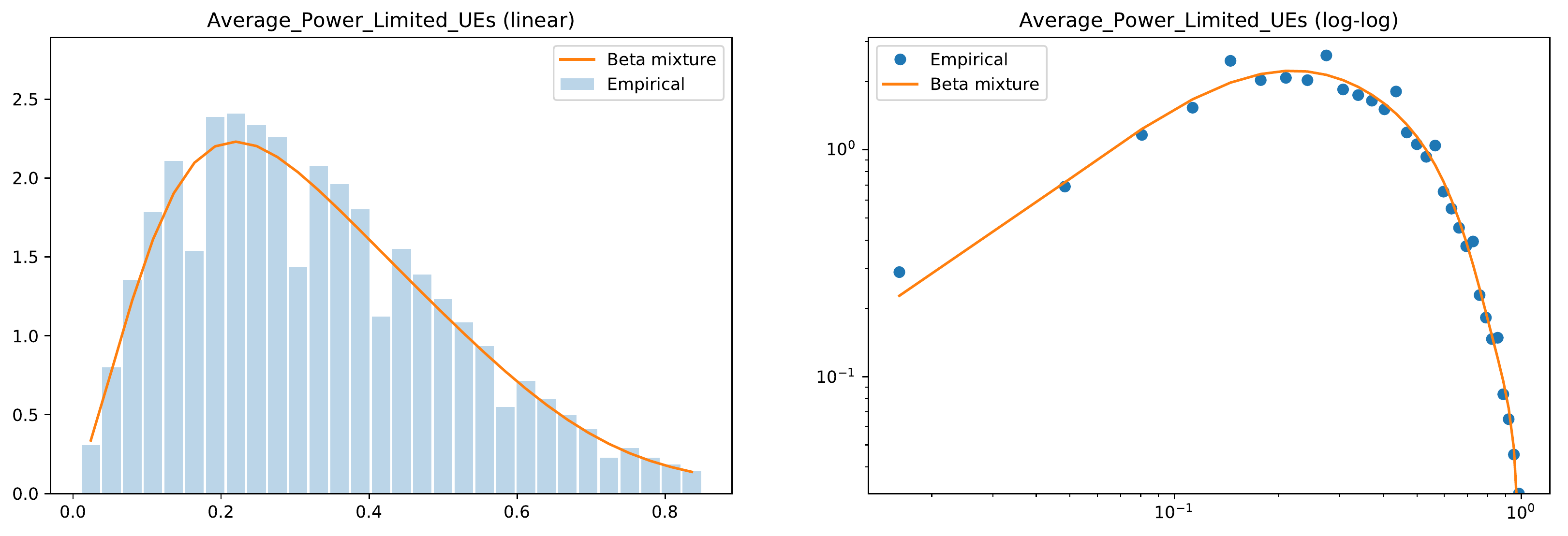}
    \includegraphics[width=0.98\linewidth]{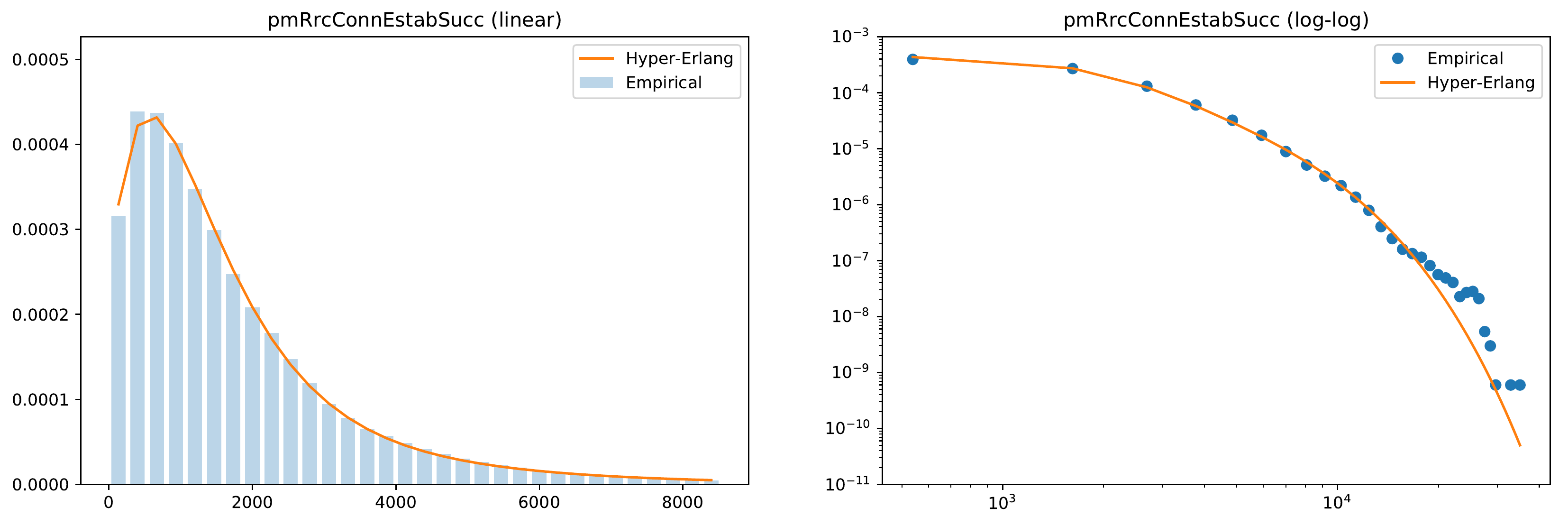}
    \caption{The marginal distributions fitting KPIs UL\_Cell\_Throughput, Average\_Power\_Limited\_UEs and pmRrcConnEstabSucc}
    \label{fig:margfit}
\end{figure}

A few examples for the marginal distribution fitting are shown in Figure \ref{fig:margfit} for some KPIs (these KPIs will be used later to demonstrate the copula and the joint distribution fitting). The results are excellent, the mixture distributions are flexible enough to capture not only the body, but also the tail of the empirical distributions. These examples represent the typical fitting accuracy, there are some better, and some worse results, too, but we managed to fit the majority of the marginal distributions with high accuracy.

\subsection{Fitting bivariate copulas}

To enable the separate fitting of the marginal distributions and the dependence structure, the two-dimensional subsets of the selected KPI pairs have been transformed to pseudo-observations. The resulting two-dimensional data has uniformly distributed marginals over $[0,1]$. The two dimensional pdf obtained this way enables to infer the dependence structure of the two random variables.

To the transformed two-dimensional data several copula families are fitted (as detailed in Section \ref{sec:fitcopula}), and the one leading to the highest likelihood is selected.
Out of the candidate copula families the Gumbel copula turned out to be the best in $3$ cases, the BB1 and the BB6 in $5$ cases each, the Student copula in $15$ cases, and the non-parametric TLL copula in $25$ cases. The copula families associated with the pairs of variables linked by the edges of the maximum information tree are also available in Figure \ref{fig:spanning}.

\begin{figure}[h]
	\centering
	\includegraphics[width=0.85\linewidth]{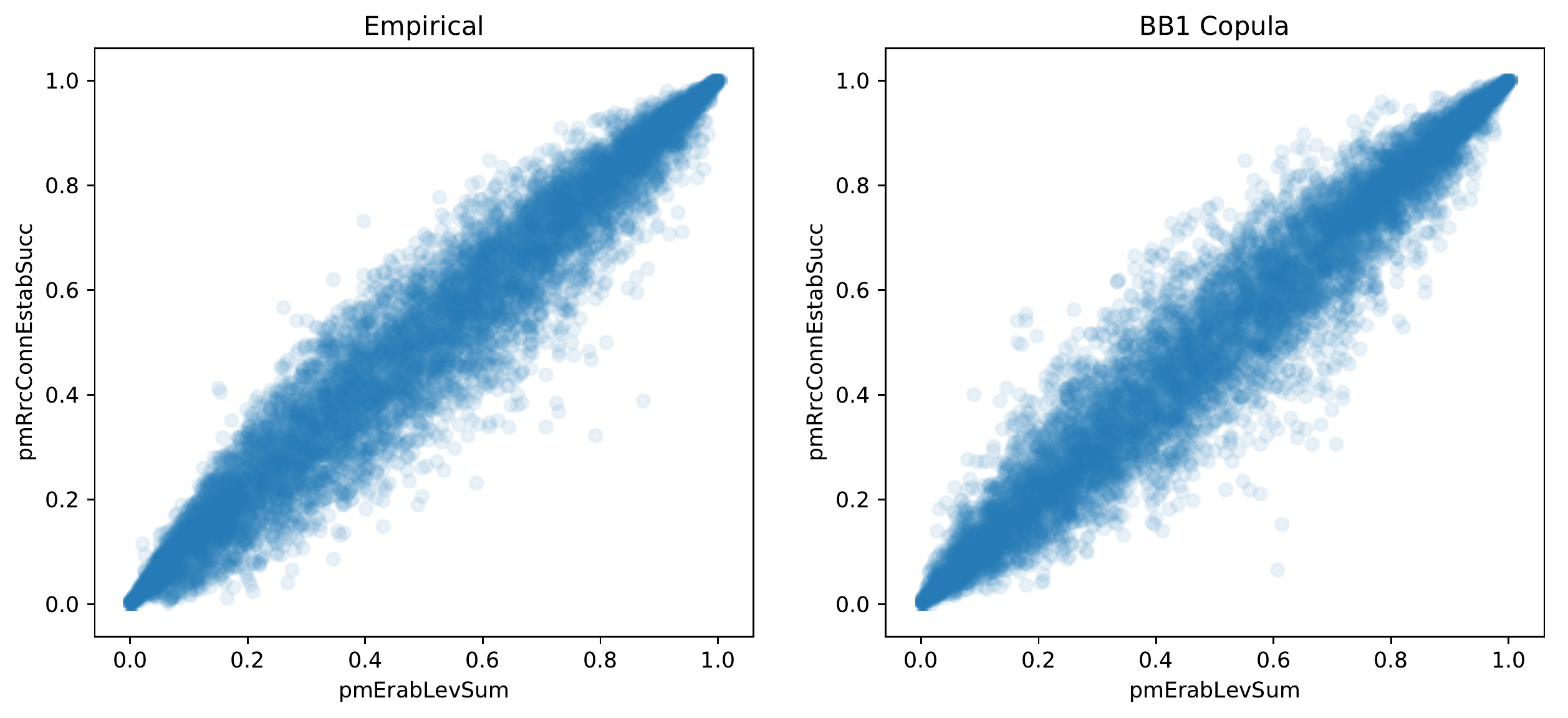}
	\includegraphics[width=0.85\linewidth]{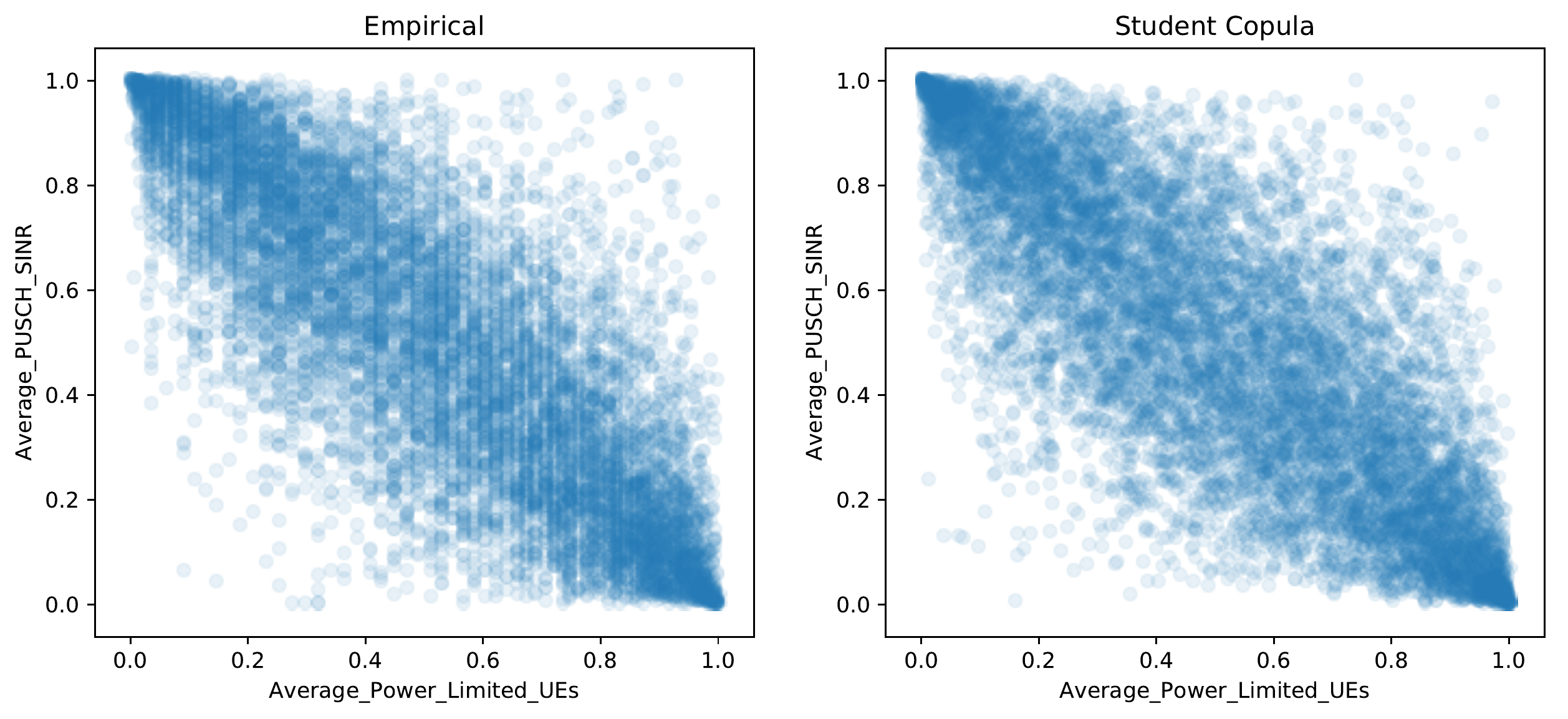}
	\includegraphics[width=0.85\linewidth]{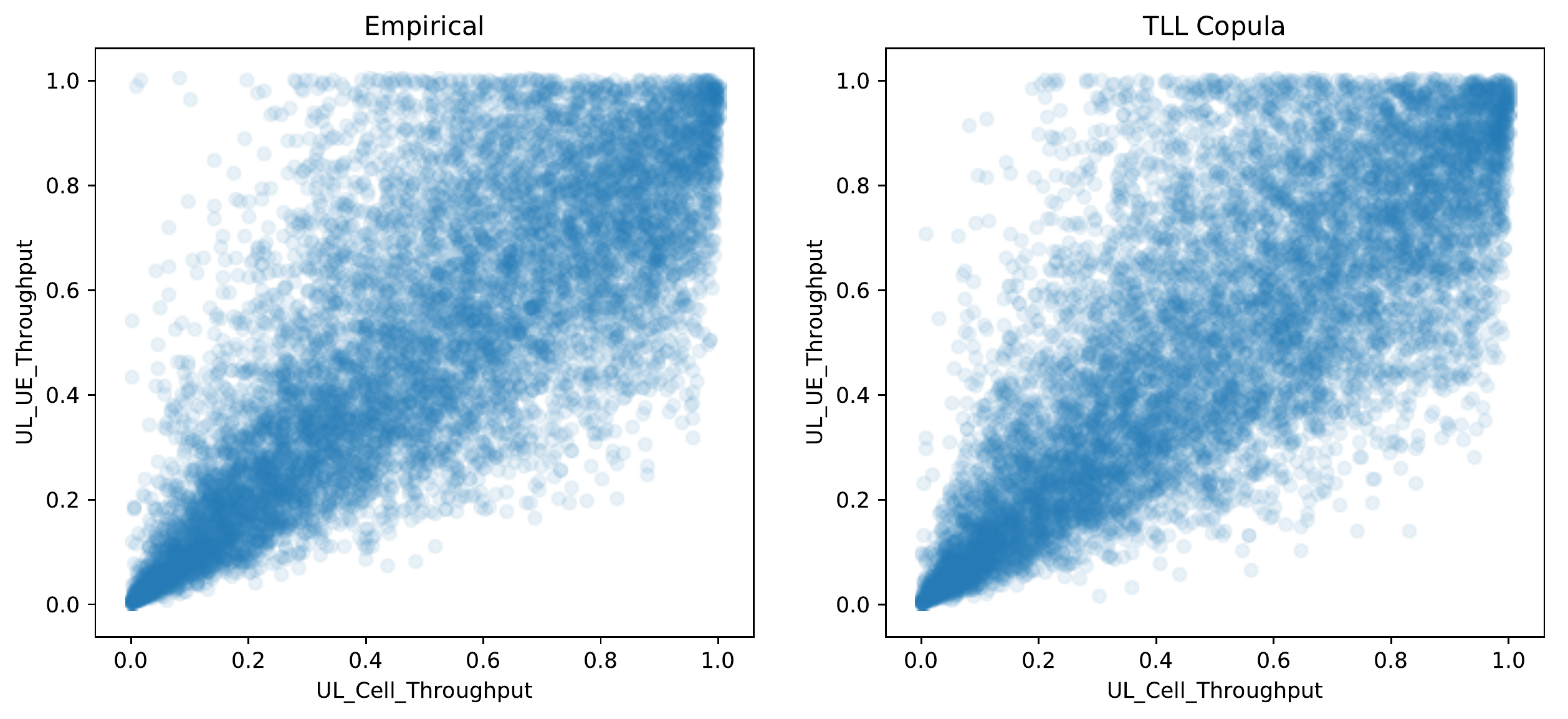}
	\caption{Examples for copula fitting. Left: empirical copula. Right: simulated copula.}
	\label{fig:copfit}
\end{figure}

Figure \ref{fig:copfit} shows 3 examples for 3 copula families comparing the transformed observations with simulated samples of the fitted copulas. In the top row the BB1 copula, in the middle the Student copula, and in the bottom row the TLL copula provides the best fit. According to the examples in the Figure (that represent typical cases), the empirical copulas occurring in our dataset can be approximated with the considered copula families with high accuracy.

\subsection{The bivariate joint distributions}

Finally, we put together the bivariate copula and the marginal probability distributions for each KPI pair, reconstructing the bivariate joint probability distributions.

\begin{figure}[h]
	\centering
	\includegraphics[width=0.85\linewidth]{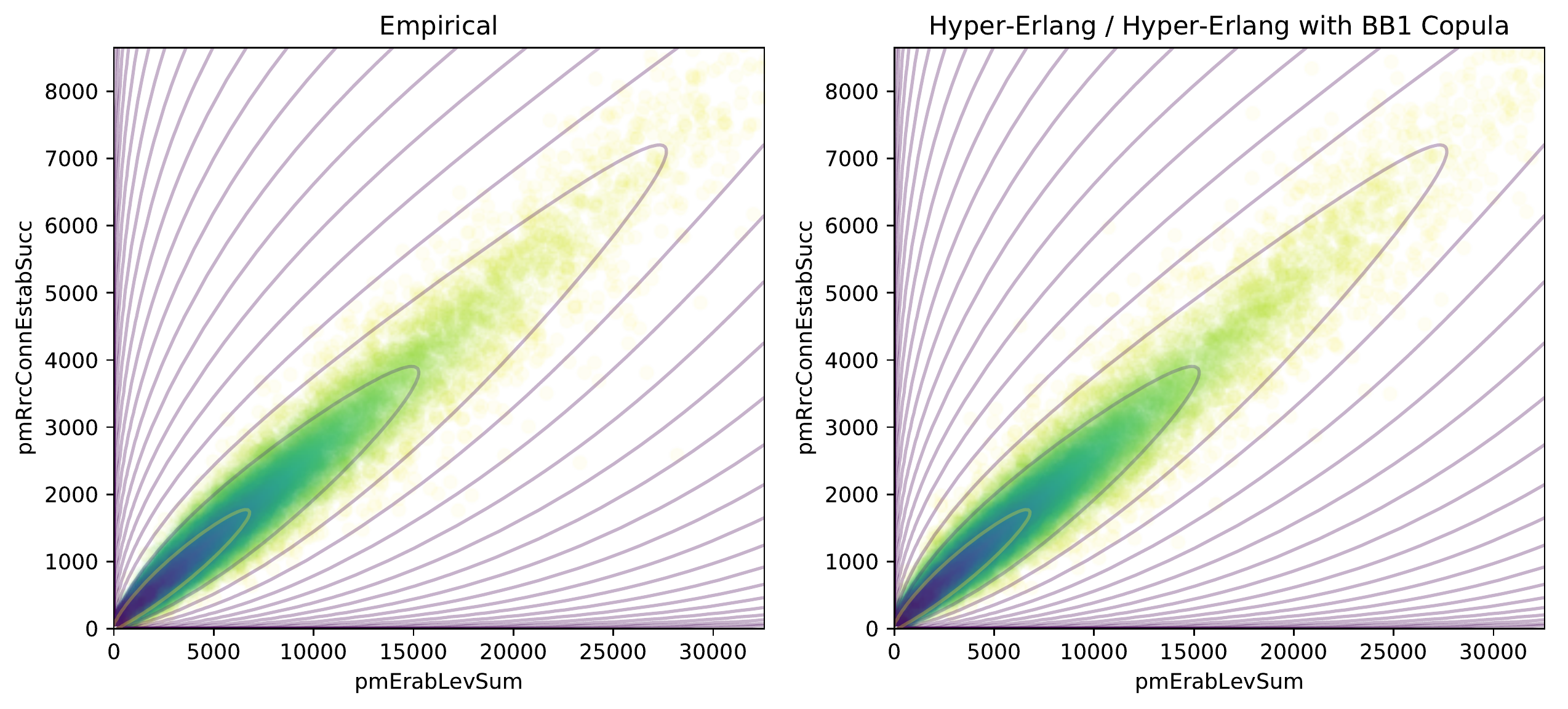}
	\includegraphics[width=0.85\linewidth]{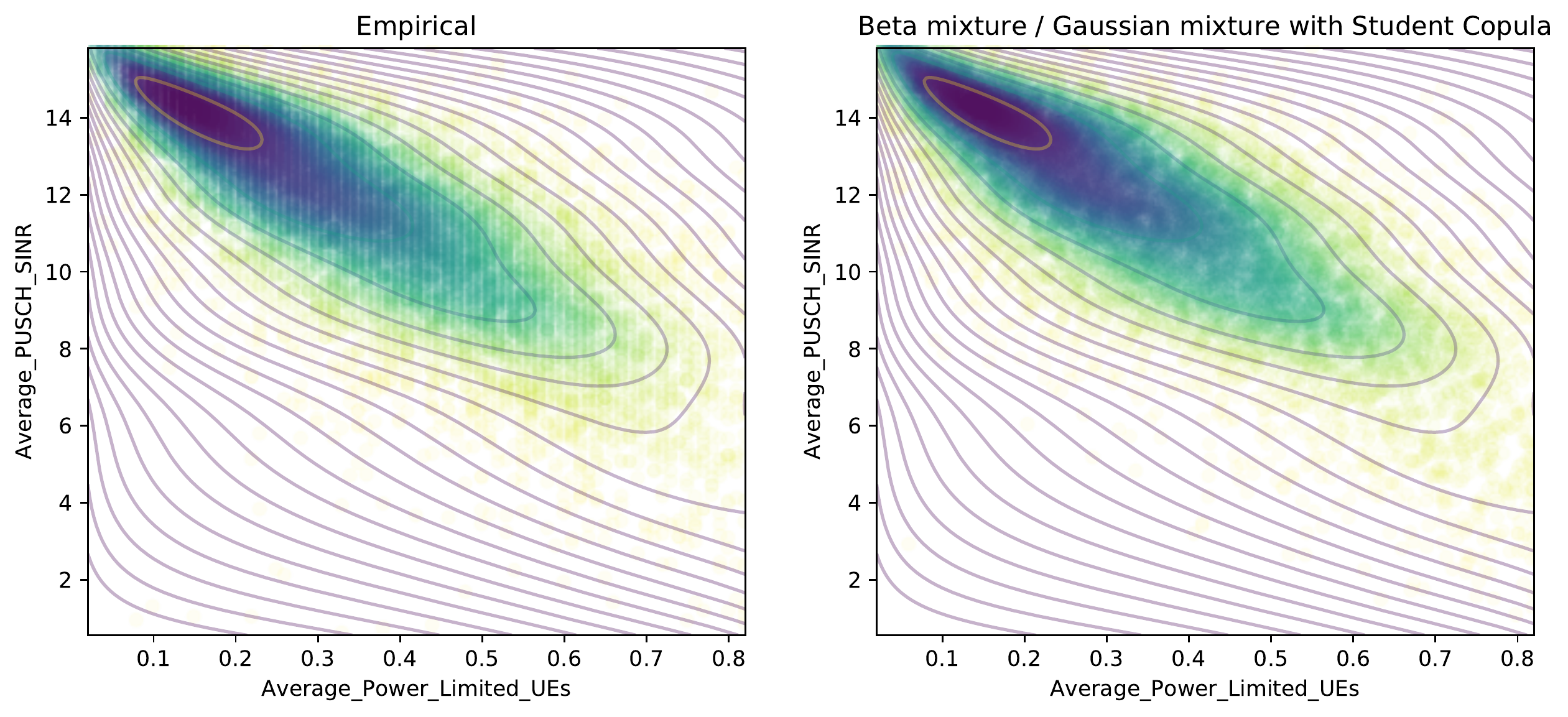}
	\includegraphics[width=0.85\linewidth]{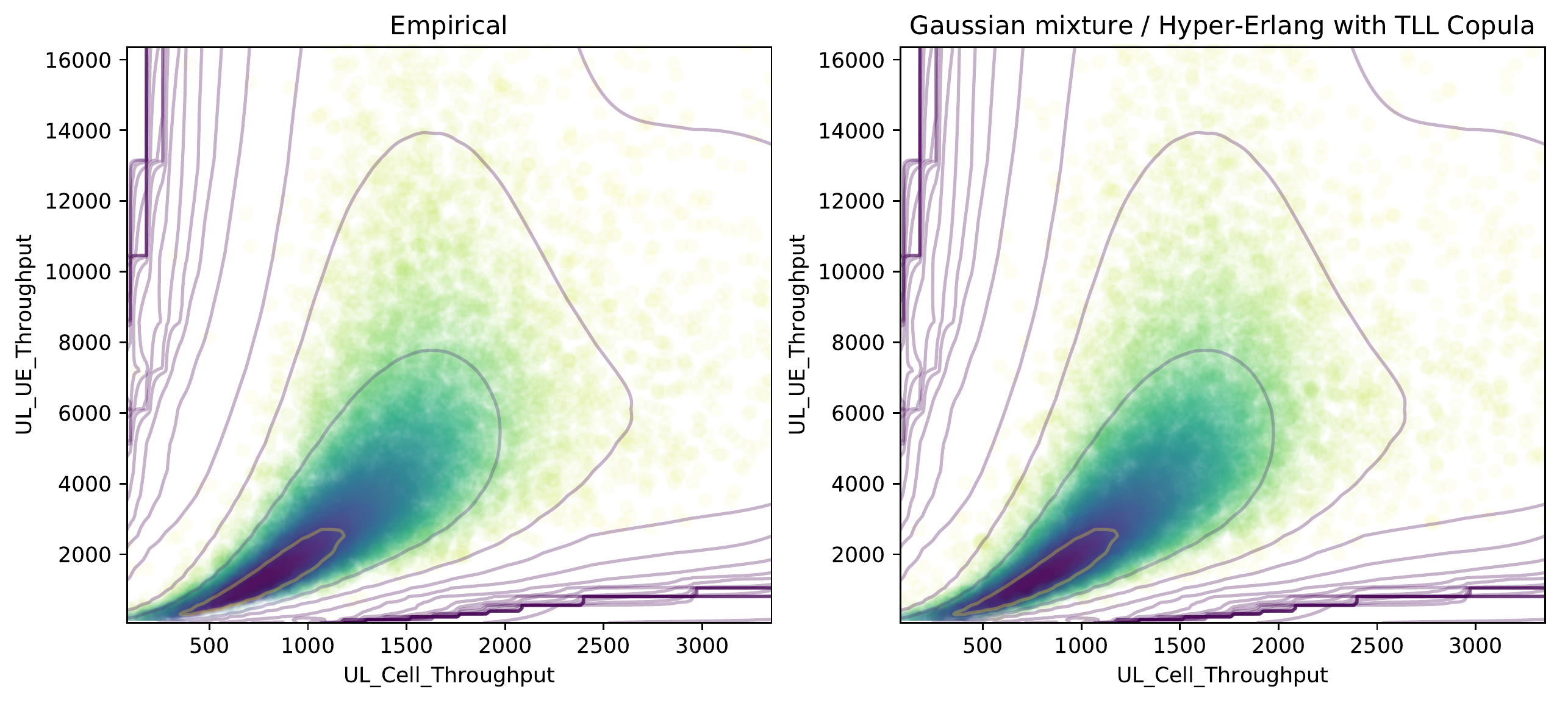}
    \caption{Reconstructing the joint distributions. Left: original data points. Right: random points simulated from the model.}
	\label{fig:bvjd}
\end{figure}

The joint distributions corresponding to Figure \ref{fig:copfit} are depicted in Figure \ref{fig:bvjd}. In the figure, the original data points are shown in the left plot and random points simulated from the modelled joint distribution are given in the right side. The purpose of the contour lines is to make the comparison easier, they are generated from the probabilistic model and are the same for both the left and right hand side plots. The color of the points reflect the anomaly score (computed according to Section \ref{sec:score}), light shades correspond to more rare (hence, more anomalous), dark shades to more typical observations. 
Observe that the joint distributions in these examples are very far from a bivariate Gaussian distribution, and are in fact difficult to fit with a mixture of bivariate Gaussians, too. However, based on the figures it is clear that the the copula-based modeling of the relevant variable pairs can be applied to this dataset successfully. 

\subsection{Anomaly tree}
To visualize the current state of the network element (i.e., an LTE cell, in our case) we introduce \emph{anomaly tree}s. 
The structure of the anomaly tree is given by the maximum information tree, and the weights of the edges are given by the anomaly scores calculated from the two dimensional marginal probability distributions. If the weights of the edges are visualized with different colors, the anomaly trees provide a useful tool for the human operators to get an overview on the current state of the network element. Form engineering point of view the anomaly tree can be considered as a high level summary of the actual anomaly state of the monitored network element. The high weight edges identify the functional part(s) of the network element that are affected by the actual anomaly. 

\begin{figure}
	\centering
	\includegraphics[width=\linewidth]{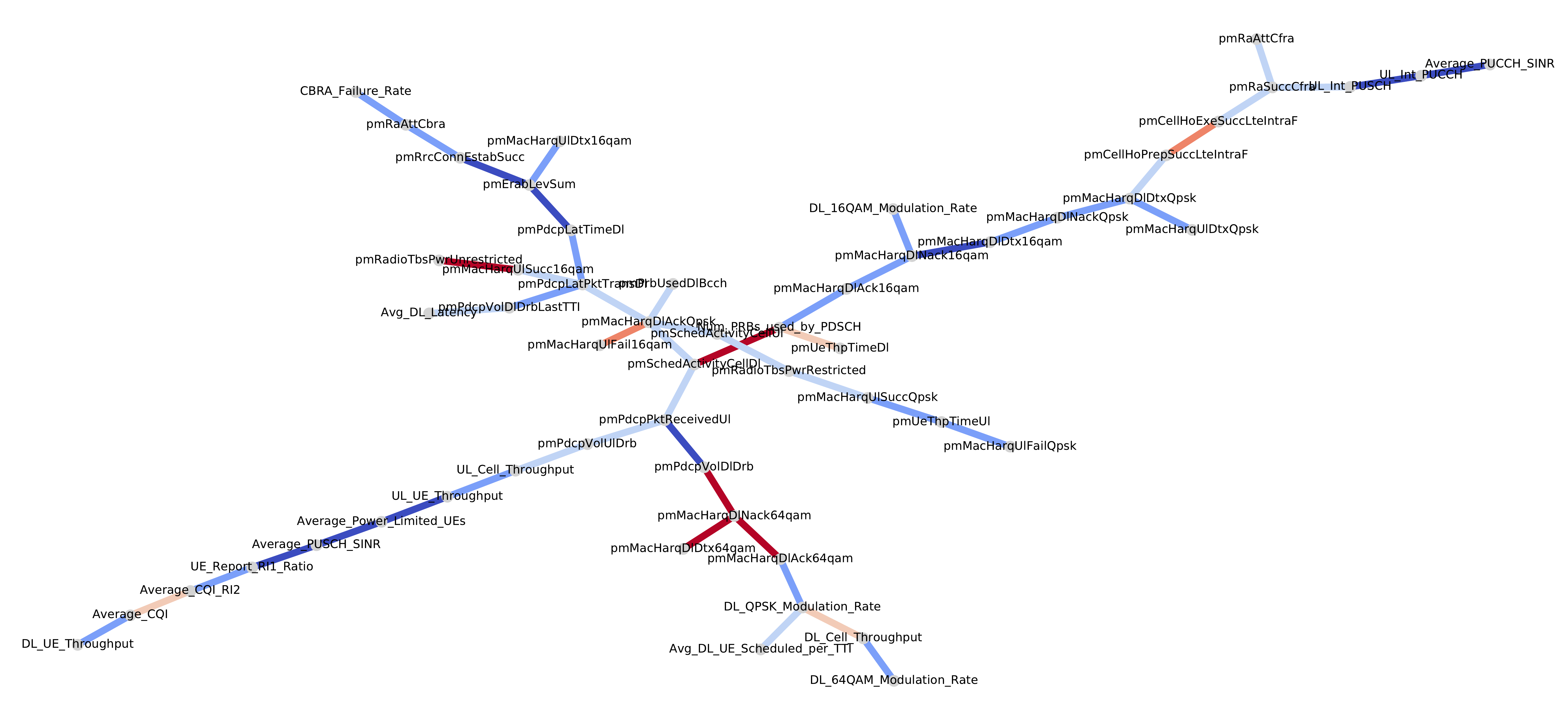}
	\caption{Anomaly trees, example 1}
	\label{fig:anotree1}
\end{figure}

\begin{figure}
	\centering
	\includegraphics[width=\linewidth]{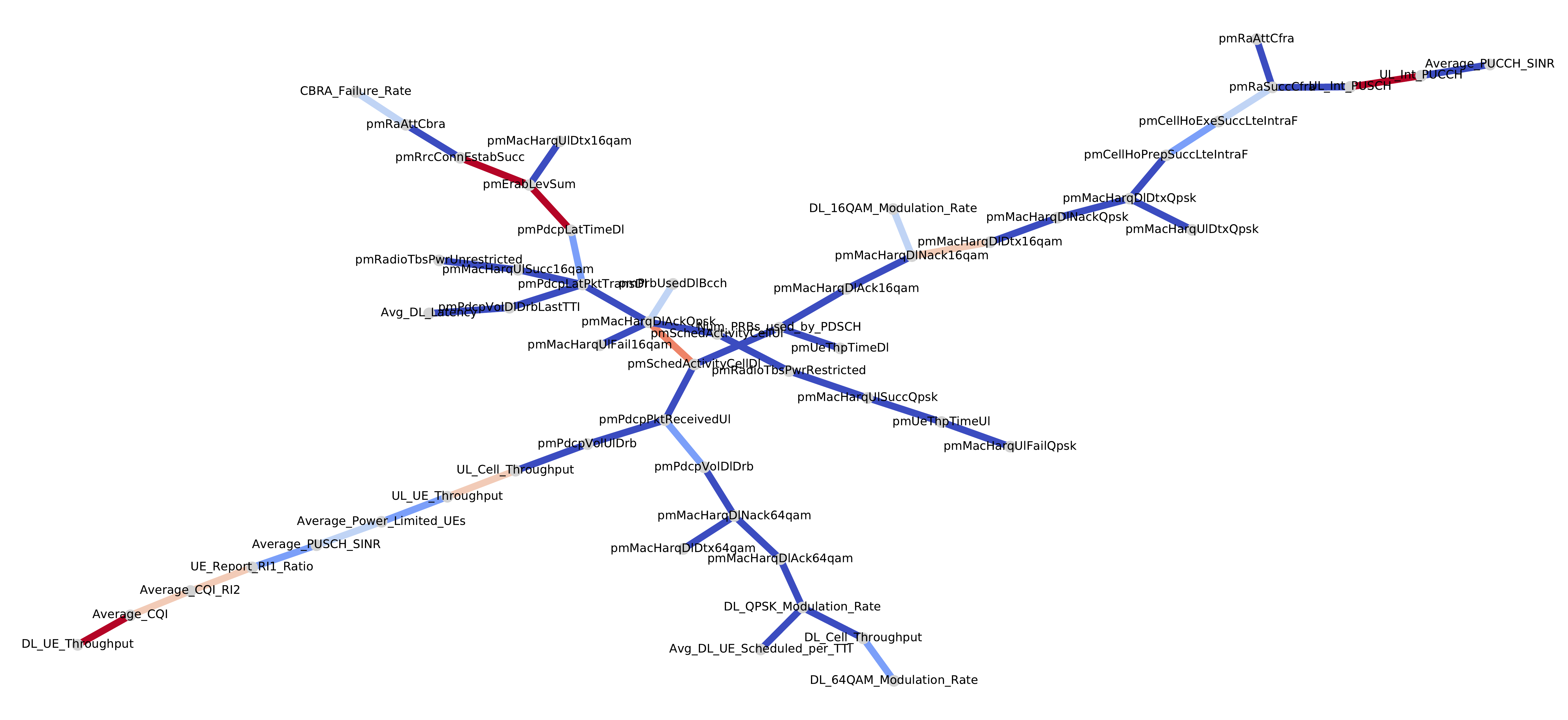}
	\caption{Anomaly trees, example 2}
	\label{fig:anotree2}
\end{figure}

Anomaly trees are illustrated in Figures \ref{fig:anotree1} and \ref{fig:anotree2}. 
In these figures, the color indicates the severity of the anomaly. For example, the red color of a link means that the anomaly score is above $0.996$. Orange shades mean lower anomaly score, and blue shades are assigned to non-anomalous relations.
Interpreting these anomaly trees requires deep domain specific knowledge, the detailed discussion of the possible reasons and the solutions is out of scope of this paper.

In Figure \ref{fig:anotree1} we can see that the number of PRBs (Physical Resource Blocks) is extremely low, compared to the number of times a user equipment is selected for transmission (affecting \texttt{pmSchedActivityCellDl} - \texttt{Num\_PRBs\_used\_by\_PDSCH} relations). Given the transport block size (\texttt{pmRadioTbsPwrUnrestricted}), the high usage of the 16-qam modulation scheme (\texttt{pmMacHarq- UlSucc16qam}) and the large number of negative acknowledgements in case of the 64-qam modulation scheme (\texttt{pmMacHarqDlNack64qam}) implies that there could be something wrong with the radio channel, possibly with the antenna.


Figure \ref{fig:anotree2} depicts an other kind of problem. In this case the channel quality is high enough, yet the user throughput is too low (according to the \texttt{Average\_CQI} - \texttt{DL\_UE\_Throughput} relation). Considering the number of e-RABs (evolved radio access bearer), the number of RRC connection establishments is high (\texttt{pmErabLevSum} - \texttt{pmRrcConnEstabSucc}), and the signalling traffic is unusually high in upload direction (\texttt{UL\_Int\_PUCCH} - \texttt{UL\_Int\_PUSCH}), which means that the reason for the bad user experience is not the quality of the channel, but the sub-optimal setting of the RRC level parameters in the cell.

The manufacturer of the base station hardware has released guidelines for  performance management and optimization, and also a troubleshooting guide, that the operators can use in order to resolve the problems detected by the presented anomaly detection algorithm.



\subsection{Comparison with alternative methods}\label{sec:altern}

In this section we demonstrate how well the presented method performs in comparison with some standard and recent anomaly detection methods.

\begin{figure}
    \centering
    \includegraphics[width=0.45\linewidth]{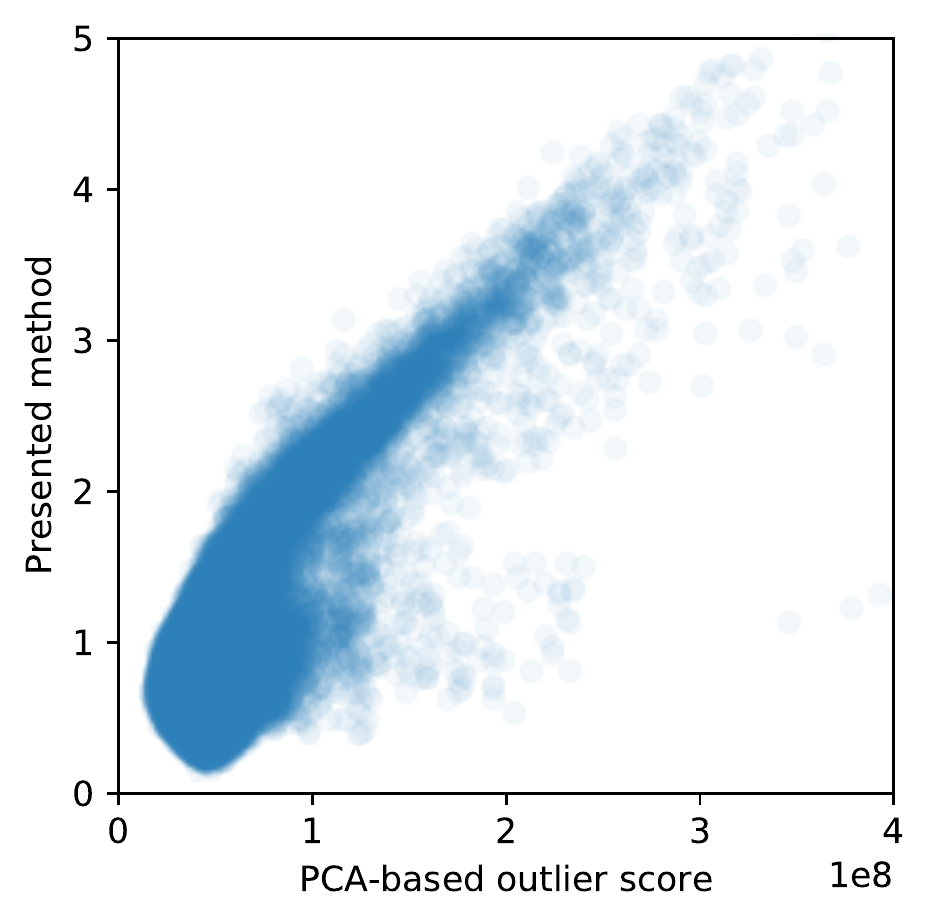}\hfill
    \includegraphics[width=0.45\linewidth]{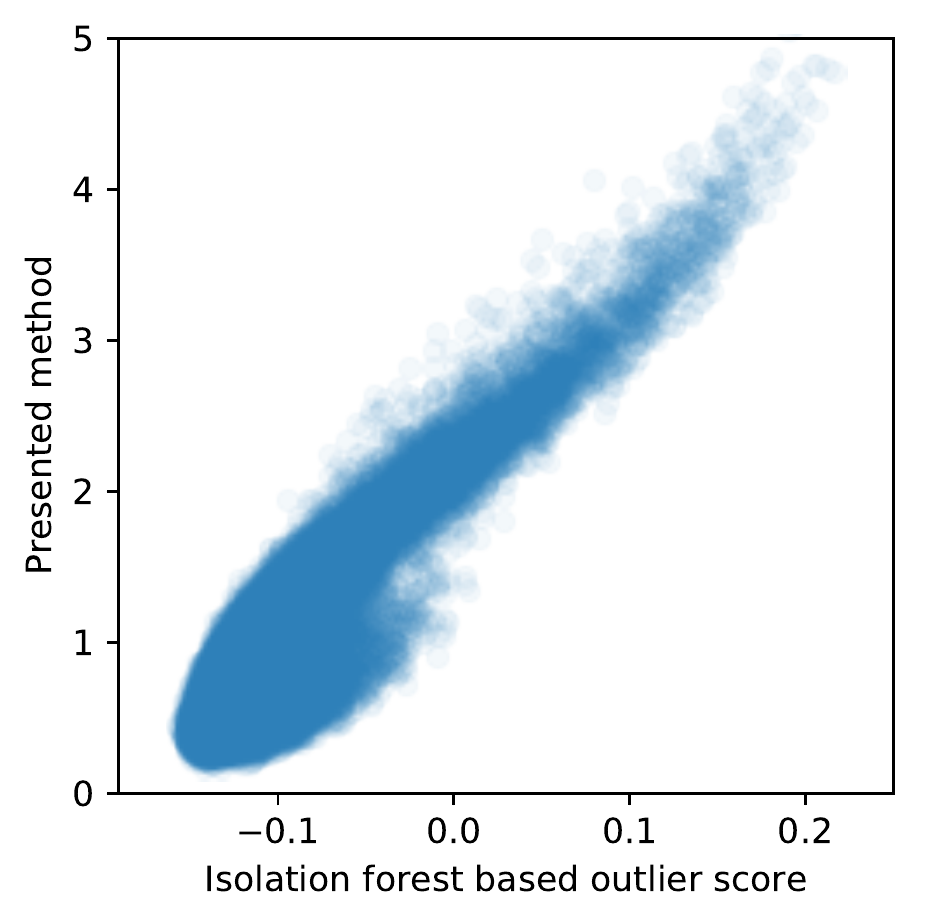} \\
    \includegraphics[width=0.45\linewidth]{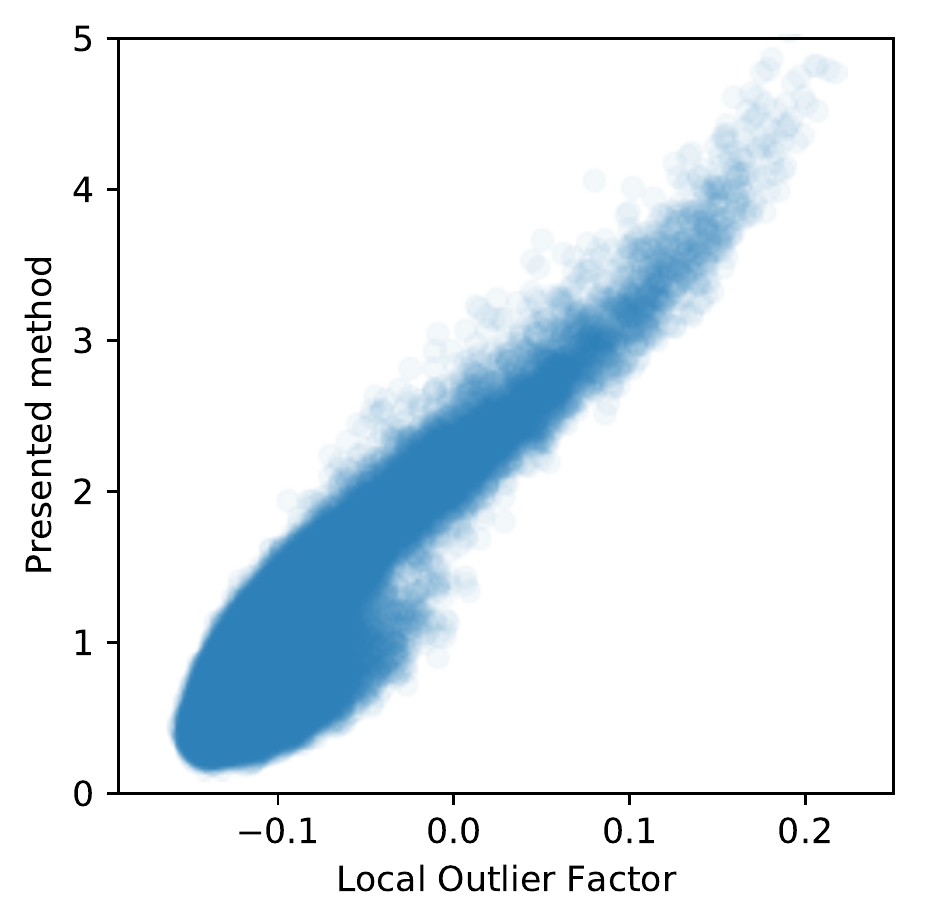}\hfill
    \includegraphics[width=0.45\linewidth]{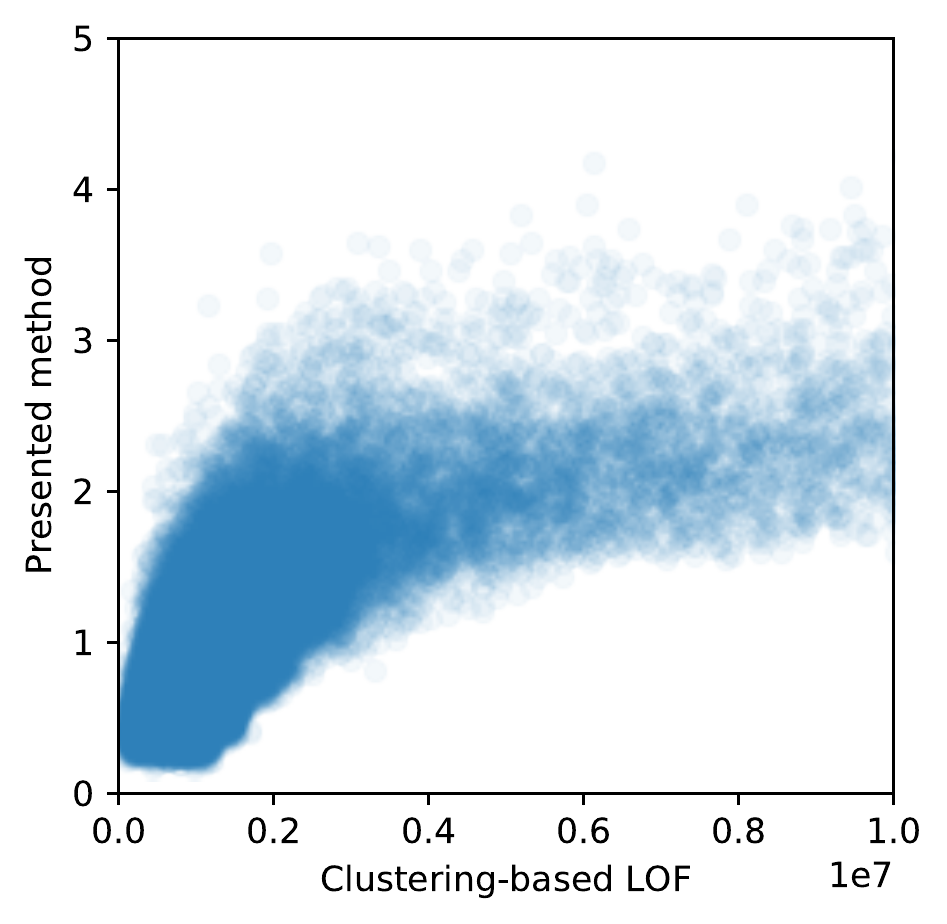} \\
    \includegraphics[width=0.45\linewidth]{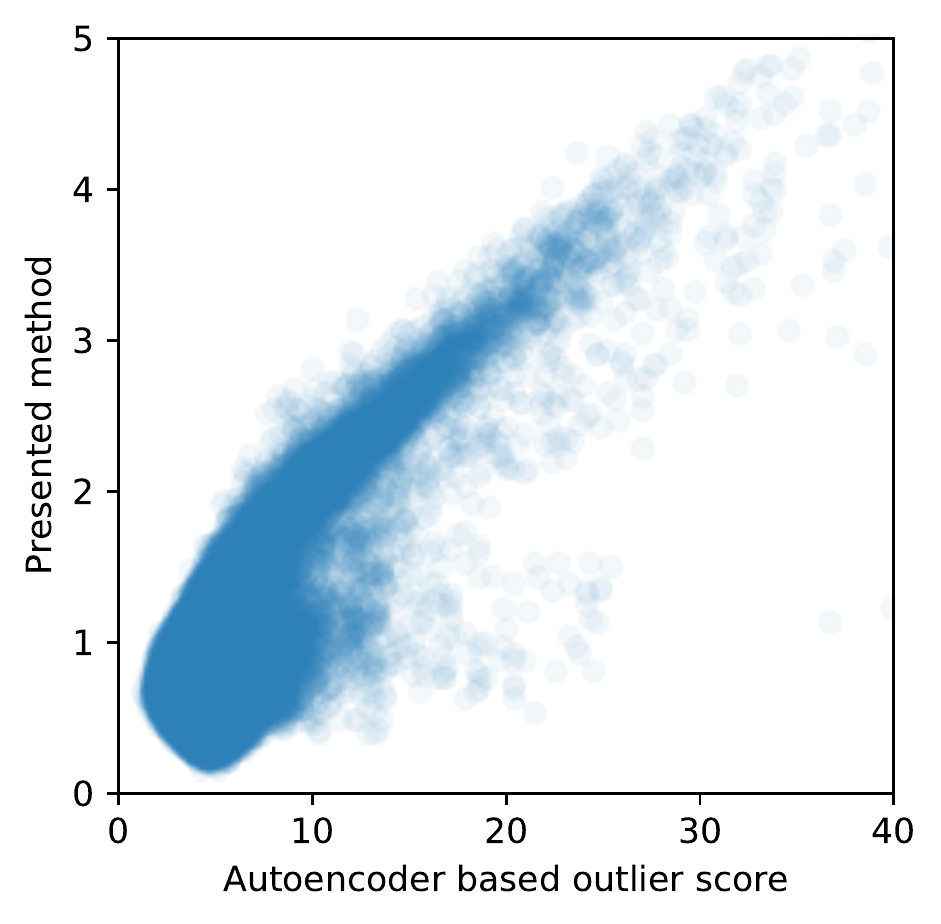}
    \caption{Comparison with alternative anomaly scoring methods}
    \label{fig:anocmp}
\end{figure}

First we ignore the most appealing capability of the presented method, that is, the anomaly localization, and extract one overall anomaly score from the the whole anomaly tree. For an observation $\mathbf{x}$ this overall anomaly score is obtained by
\begin{align}
    \tilde{\mathcal{A}}(\mathbf{x}) = \frac{1}{d-1}\sum_{e\in T} - \log(1-\hat{\mathcal{A}}(\mathbf{x_e})),
\end{align}
where $\mathbf{x_e}$ is the two-dimensional sample containing only those two components of $\mathbf{x}$ that correspond to edge $e$ of the tree $T$. The role of the logarithm is to adjust the scale of the anomaly score such that only really rare observations will get high scores. Hence, after the logarithm re-scaling, we take the average anomaly scores of the anomaly tree.

Then, we compare this overall anomaly score to anomaly scores obtained by alternative methods. It is important to note that all the alternative methods we tested work on complete data only, and the LTE network data set contains only $66,000$ complete samples out of the $1,500,000$ observations. Thus, in the rest of this section all comparisons are based on this very restricted, but complete data set.

To obtain anomaly scores from alternative procedures we have used the Python Toolbox for Scalable Outlier Detection (PyOD, \cite{zhao2019pyod}). Several procedures failed to provide a result in a reasonable time, or gave constant zero anomaly score for all samples (including the angle-based method \cite{kriegel2008angle}). For a comparison, Figure \ref{fig:anocmp} shows the outlier scores of the presented methods against the scores of all the procedures that returned valid result, namely PCA-based outlier score~\cite{shyu2003novel}, Isolation forest~\cite{liu2008isolation}, Local Outlier Factor~\cite{breunig2000lof}, Clustering-based LOF~\cite{he2003discovering} and Autoencoder based outlier score~\cite{aggarwal2015outlier}. For this study we have used the implementation available at \url{https://github.com/yzhao062/pyod} with the default parameters, as listed by Table \ref{tab:defautls}.

From Figure \ref{fig:anocmp} it can be seen that our method assigns high anomaly score to those observations that are found anomalous by alternative methods as well. Additionally, our method can provide information on the location of the anomaly and can operate with missing data as well. The fact that in a real data set missing data occurs frequently (possibly due to unreliable sensors) justifies the need for anomaly detection procedures that tolerate missing data. 

\begin{table}
\centering
    \begin{tabular}{c|c}
        Anomaly detection method & Parameters \\
        \hline
        PCA based method & use all components \\
        Isolation forest & n\_estimators=$100$ \\
        Local Outlier Factor & n\_neighbors=$20$ \\
        Clustering-based LOF & n\_clusters=$8$, $\alpha=0.9$, $\beta=5$ \\
        Autoencoder based method & neurons=$[64, 32, 32, 64]$, activation=relu, epochs=20
    \end{tabular}
    \caption{Parameters of the methods involved in the comparison}
    \label{tab:defautls}
\end{table}

\section{Evaluation of the method on a synthetic data set}\label{sec:synth}

We have prepared an other numerical study to compare the presented copula based method to alternative anomaly detection methods published in the literature.

\begin{figure}[h]
	\centering
	\includegraphics[width=0.85\linewidth]{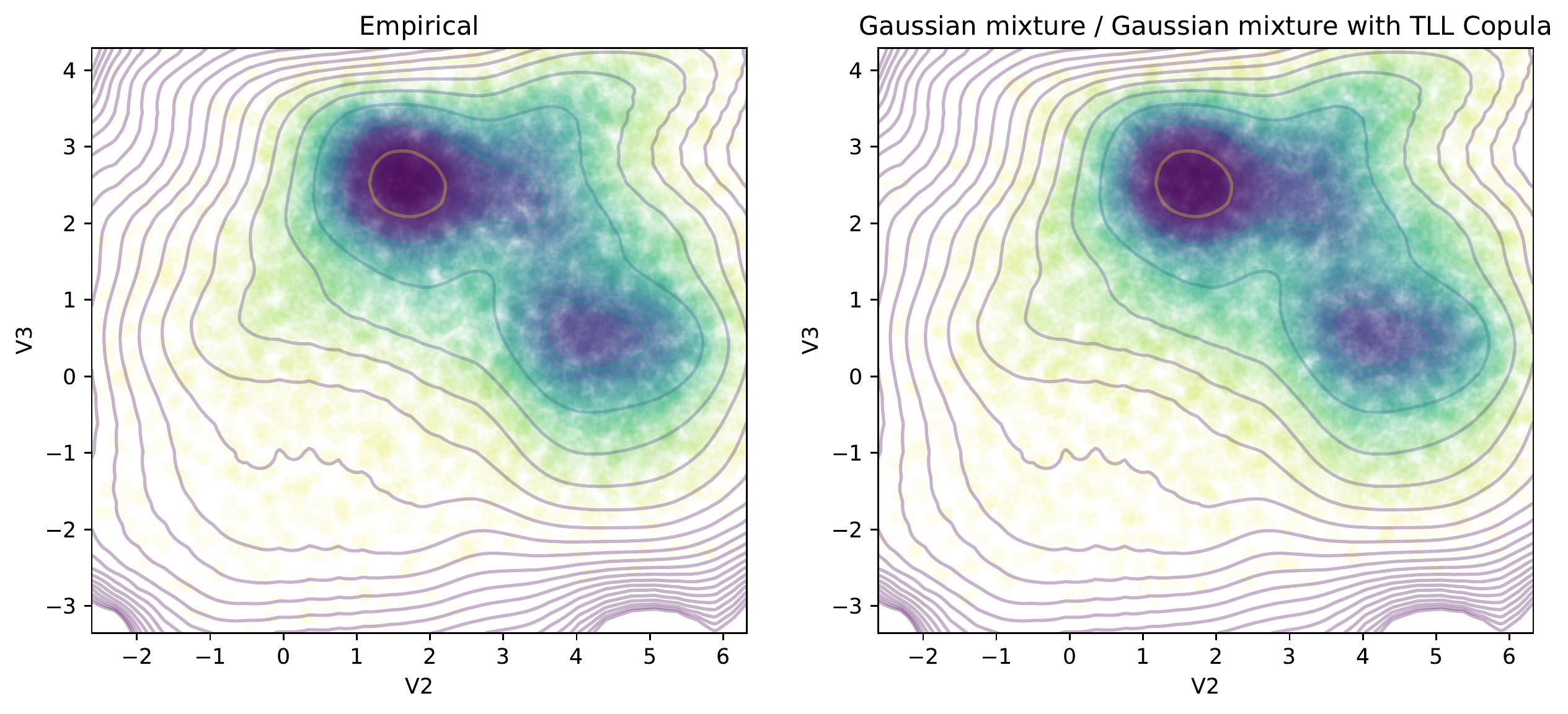}
    \caption{One of the bi-variate joint distributions in the synthetic example. Left: original data points. Right: random points simulated from the model.}
	\label{fig:bvjd-syn}
\end{figure}

In this study we generate one million samples from a known distribution, and investigate how the anomaly scores correlate with the probability densities. For this purpose, we have defined a mixture of $5$-dimensional normal distribution consisting of $4$ components. The exact parameters of the distribution are provided in Appendix \ref{sec:dparams}. This choice may seem to be unfair, since our method is able to capture its marginals with high accuracy, but it turns out that this distribution has a very complicated correlation structure, that is equally difficult for all procedures involved in the comparison. As shown in Figure \ref{fig:bvjd-syn}, a bi-variate joint distribution is multi-modal, our model can still capture its characteristic rather well.

The procedures involved in the comparison are the same as in Section \ref{sec:altern}, namely the PCA based, the isolation forest based, the Local Outlier Factor based, the Clustering-based LOF and the Autoencoder based methods. Since the samples are generated from a know distribution, we can calculate the Kendall rank correlation coefficient between the anomaly scores returned by these methods and the values of the probability density function. The results are summarized in Table \ref{tab:ascorr}. The correlations are negative, since higher anomaly scores are assigned to observations having lower densities. According to the results, the presented copula based method performs the best for this particular data set.

\begin{table}[]
    \centering
    \begin{tabular}{c|c}
        Anomaly detection method & Kendall's tau \\ 
        \hline
        PCA based scoring & -0.337 \\
        Isolation forest & -0.526 \\
        Local Outlier Factor  & -0.4194 \\
        Clustering-based LOF  & -0.5892 \\
        Autoencoder based method & -0.3815 \\
        \textbf{Presented method}  & \textbf{-0.6275}
    \end{tabular}
    \caption{The correlation between the anomaly scores and the probability densities}
    \label{tab:ascorr}
\end{table}

We emphasize again, that this does not mean that the copula based method is always better than other methods, it just means that for data sets consisting of a high number of continuous variables it performs favorably. The alternative methods still have their strengths, when the data set is smaller, has categorical features and probability masses.

\begin{figure}
    \centering
    \includegraphics[width=0.45\linewidth]{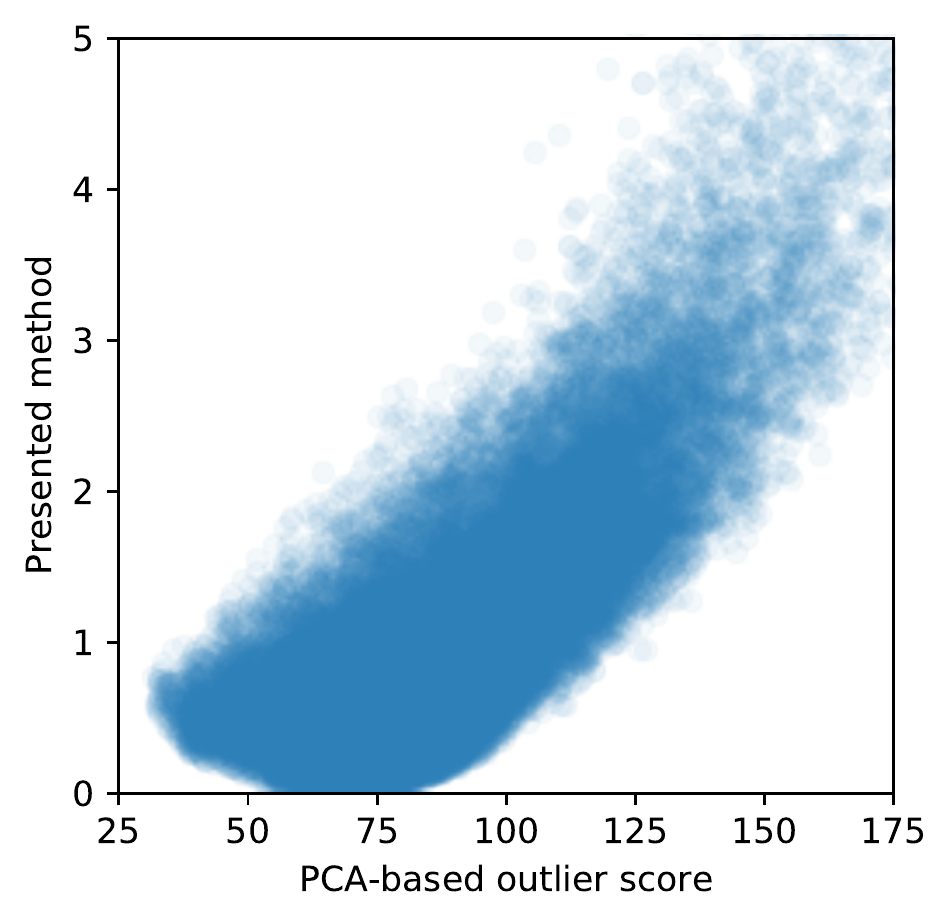}\hfill
    \includegraphics[width=0.45\linewidth]{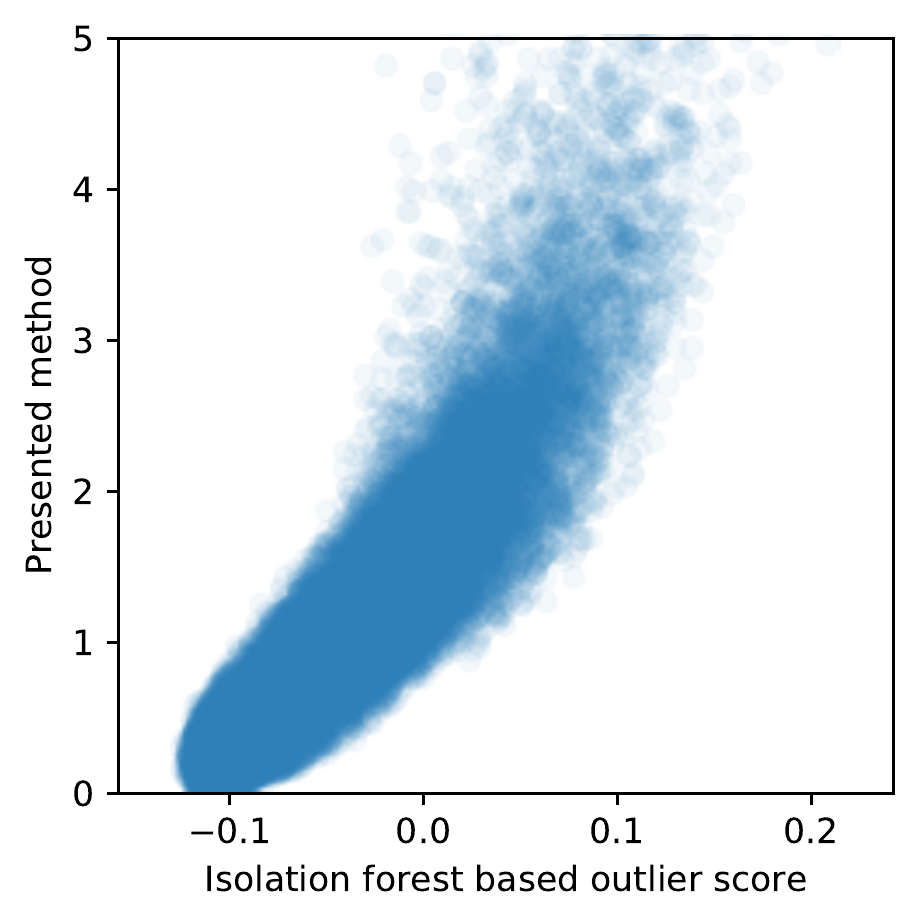} \\
    \includegraphics[width=0.45\linewidth]{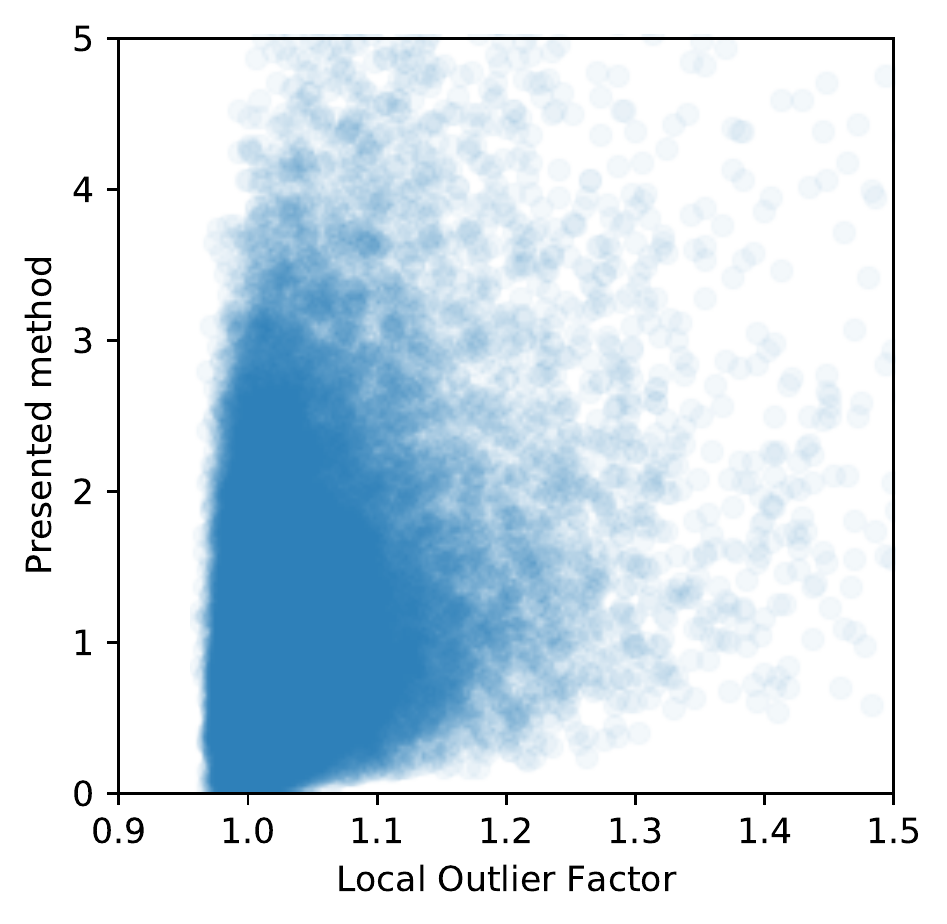}\hfill
    \includegraphics[width=0.45\linewidth]{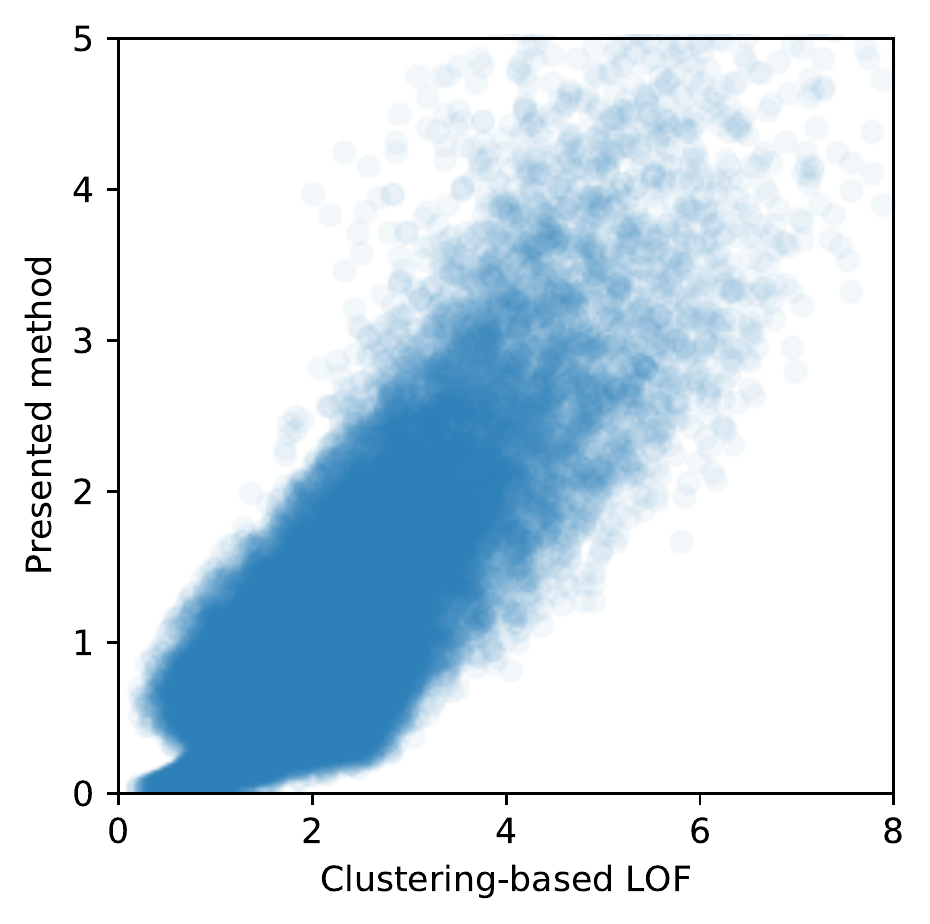} \\
    \includegraphics[width=0.45\linewidth]{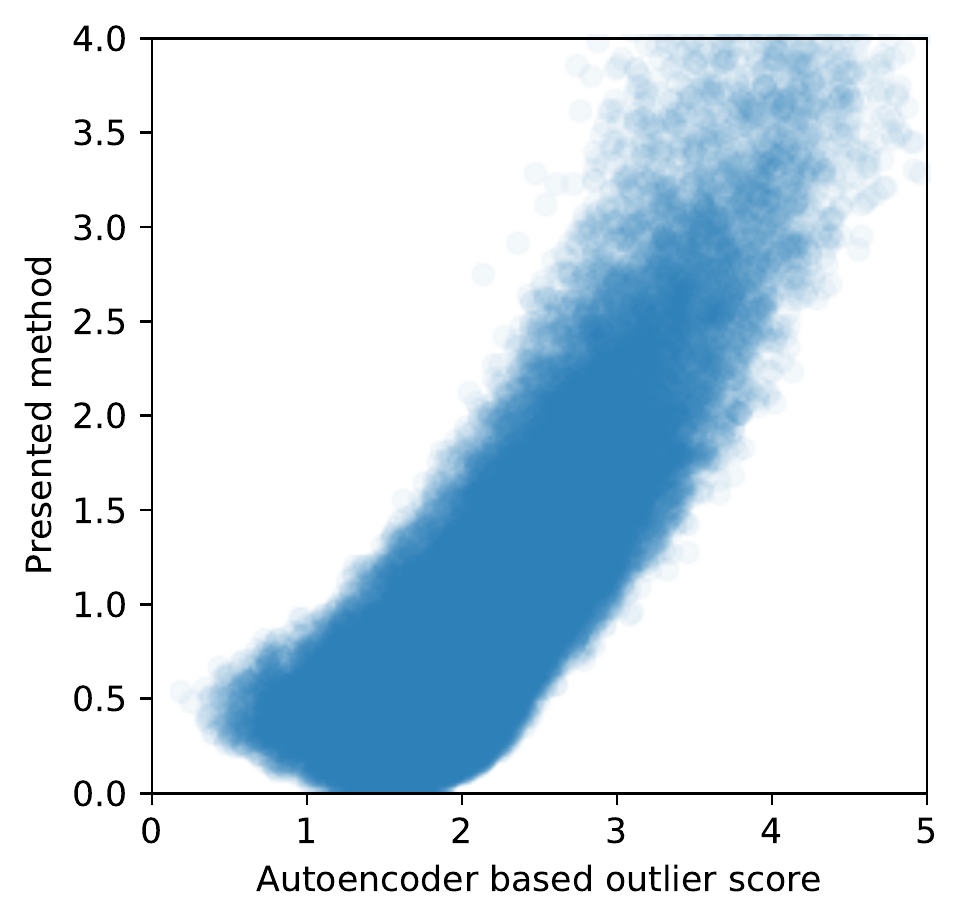}
    \caption{Comparison with alternative anomaly scoring methods on the synthetic data set}
    \label{fig:anocmp-syn}
\end{figure}

For completeness, we include the plots depicting the correlation between the copula based anomaly scores and the anomaly scores of the alternative methods, see Figure \ref{fig:anocmp-syn}.

\section{Conclusion and discussion}

In this paper, we presented a novel copula-based anomaly detection technique for high-dimensional data that does not only assign an anomaly score to the observations but it also localizes the reason of the anomaly. The proposed approach relies on the modeling of the multivariate probability distribution associated with the instances. Our procedure is able to handle large-scale, high-dimensional data since the intractable high-dimensional problem is broken into smaller tractable ones by using two dimensional projections (bivariate marginal distributions) of the joint probability distribution in such a way that it retains maximum information and reduces redundancy. Since rare events occur in the tails of the probability distribution, copulas were used to model the bivariate marginals. Another advantage of the copula approach that the univariate marginal probability distributions and the bivariate copulas can be fitted separately. 

Besides an overall anomaly score, our approach also reports individual anomaly scores for the selected variable pairs that can be illustrated on an anomaly tree that enables the users to get an overview on the current state of the system, and to observe it evolving in time. We also illustrated our method on a real-world telecommunication problem. Our findings were reconfirmed by real network operators. Moreover, we compared the anomaly score of the proposed approach with other standard anomaly detection methods. We can conclude that the overall anomaly score returned by our approach is similar to the scores returned by other methods. On the other hand, our method also provides support to localize the anomaly, can be parallelized, can cope with a large number of high-dimensional observations and is also capable of handling missing data.

Although our proposed technique complements the existing approaches in several aspects, it also has its limitations. The most important limitation of the proposed approach is that it only works well with large scale continuous data. Categorical or discrete variables must be omitted since our method is not able to handle them (because the transformation to pseudo-observations does not result in a uniform distribution). This limitation also makes it troublesome to carry out a comprehensive comparison of the performance of our proposed approach with other alternative methods. Therefore, the method introduced in this paper supplements and does not substitute the existing techniques.

\begin{acks}
The research reported in this paper was supported by the Higher Education
Excellence Program of the Ministry of Human Capacities in the frame of Artificial Intelligence research area of Budapest University of Technology
(BME FIKP-MI/SC). The publication is also supported by the EFOP-3.6.2-16-2017-00015 project entitled "Deepening the activities of HU-MATHS-IN, the Hungarian Service Network for Mathematics in Industry and Innovations". The project has been supported by the European Union, co-financed by the European Social Fund. The research reported in this paper has been also supported by the National Research, Development and Innovation Fund (TUDFO/51757/2019-ITM, Thematic Excellence Program). 
G\'abor Horv\'ath is supported by the OTKA K-123914 project. Roland Molontay is supported by NKFIH K123782 research grant and by MTA-BME Research Group.
\end{acks}

\bibliographystyle{ACM-Reference-Format}
\bibliography{our-bibliography}

\appendix
\section{The parameters of the distribution used in Section \ref{sec:synth}} \label{sec:dparams}

To generate the syntetic data set for Section \ref{sec:synth} we have defined a mixture of $5$-dimensional normal distribution with $4$ components, having density function
\begin{align*}
    f(\mathbf{x}) = \sum_{i=1}^4 p_i \frac{1}{\sqrt{(2\pi)^5 |\Sigma_i|}} e^{-(\mathbf{x}-\mathbf{\mu}_i)^T\mathbf{\Sigma_i}^{-1}(\mathbf{x}-\mathbf{\mu}_i)/2}.
\end{align*}

The mean values of the components are given by 
\begin{alignat*}{2}
    \mathbf{\mu}_1 = \begin{bmatrix}2.9 & 4.3 & 0.6 & 4.0 & 4.4\end{bmatrix}^T, && \mathbf{\mu}_2 = \begin{bmatrix}4.0 & 1.5 & 2.7 & 0.7 & 4.5\end{bmatrix}^T, \\
    \mathbf{\mu}_3 = \begin{bmatrix}3.8 & 1.1 & 0.5 & 2.8 & 0.4\end{bmatrix}^T, && \mathbf{\mu}_4 = \begin{bmatrix}3.7 & 2.5 & 2.5 & 2.0 & 0.7\end{bmatrix}^T,
\end{alignat*}
and the mixing probabilities are $p_1=0.4, p_2=0.2, p_3=0.1, p_4=0.3$.

Obtaining the covariance matrices is a bit more involved. Our goal is to define the eigenvalues of the covariance matrices, and rotate them in a random way with unitary matrices. The eigenvalues of the covariance matrices $i=1,\dots,4$, denoted by $\mathbf{v}_i$, are given by
\begin{alignat*}{2}
    \mathbf{v}_1&=\begin{bmatrix}0.25 & 0.5 & 1 & 2 & 4\end{bmatrix}, &
    \mathbf{v}_2&=\begin{bmatrix}0.15 & 0.25 & 0.35 & 0.5 & 1\end{bmatrix}, \\
    \mathbf{v}_3&=\begin{bmatrix}1 & 2 & 3 & 4 & 5\end{bmatrix}, &
    \mathbf{v}_4&=\begin{bmatrix}0.15 & 0.25 & 1 & 4 & 5\end{bmatrix}. \\
\end{alignat*}

Before providing the covariance matrices, let us introduce the following auxiliary matrices:
\begin{alignat*}{2}
\mathbf{W}_1 &= \begin{bmatrix}
0.6 & 0.1& 0.9& 0.9& 0.4\\ 0.8& 0.7& 0.2& 0.8& 0.6\\ 0.8& 0.1 & 0.5& 0.3& 0.4\\ 0.8& 0.6& 0.7& 0.7& 0.8\\ 0.6& 0.6& 0.8& 0.7& 0.7
\end{bmatrix}, &
\mathbf{W}_2 &= \begin{bmatrix}
0.2& 0.4& 0.5& 0.6& 0.1\\ 0.8& 0.7& 0.5& 0.0& 0.8\\ 0.9& 0.2& 0.2& 0.8& 0.8\\ 0.3& 0.2& 0.5& 0.3& 0.7\\ 0.8& 0.8& 0.8& 0.3& 0.5
\end{bmatrix}, \\
\mathbf{W}_3 &= \begin{bmatrix}
0.1& 0.3& 0.0 & 0.9& 0.8\\ 0.5& 0.7& 0.5& 0.6& 0.1\\ 0.0 & 0.3& 0.2& 0.3& 0.2\\ 0.5& 0.7& 0.8& 0.5& 0.6\\ 0.3& 0.1& 0.9& 0.6& 0.1
\end{bmatrix}, &
\mathbf{W}_4 &= \begin{bmatrix}
0.4& 0.4& 0.3& 0.2& 0.5\\ 0.7& 0.4& 0.3& 0.4& 0.6\\ 0.0 & 0.8& 0.3& 0.9& 0.8\\ 0.9& 0.9& 0.1& 0.8& 0.1\\ 0.1& 0.3& 0.3& 0.9& 0.1
\end{bmatrix}.
\end{alignat*}

Applying QR decomposition on matrices $\mathbf{W}_i,i=1,\ldots,4$ gives orthogonal matrices $\mathbf{Q}_i,i=1,\ldots,4$, from which the covariance matrices are obtained by  $\mathbf{C}_i=\mathbf{Q}_i\, \mathrm{diag}(\mathbf{v}_i)\,\mathbf{Q}_i^T $.

\end{document}